\documentclass{article} 
\usepackage[numbers,sort&compress]{natbib}
\usepackage{pallas}
\usepackage{microtype}
\renewenvironment{abstract}{\begin{justify}\authorFont\setstretch{1.1}\setlength{\parindent}{0pt}}{\end{justify}}
\makeatletter
\renewcommand{\maketitle}{%
  {\centering\LARGE\titleFont\textcolor{titleblue}{\@title}\par}%
  \vspace{0.5em}%
  {\centering\normalsize\authorFont\@author\par}%
  \vspace{0.5em}}
\makeatother

\usepackage{amsmath,amsfonts,bm}

\def\eqref#1{equation~\ref{#1}}

\def\1{\bm{1}}

\DeclareMathAlphabet{\mathsfit}{\encodingdefault}{\sfdefault}{m}{sl}
\SetMathAlphabet{\mathsfit}{bold}{\encodingdefault}{\sfdefault}{bx}{n}

\usepackage{hyperref}
\usepackage{url}
\usepackage{booktabs}
\usepackage{graphicx}
\usepackage{amsmath}
\usepackage{amssymb}
\usepackage{multirow}
\usepackage{float}
\usepackage{algorithm}
\usepackage{algpseudocode}

\graphicspath{{figs/}}

\usepackage{xcolor}
\usepackage{etoc}
\usepackage{enumitem}
\usepackage{wrapfig}

\definecolor{linkblue}{HTML}{0060E0}

\hypersetup{
    colorlinks=true,
    linkcolor=titleblue,
    citecolor=titleblue,
    urlcolor=linkblue,
    pdftitle={Towards Continual Learning: Mitigating Forgetting With Subspace Tuning}
}

\usepackage{cleveref}

\crefname{section}{Sec.}{Secs.}
\Crefname{section}{Sec.}{Secs.}

\crefname{subsection}{Sec.}{Secs.}
\Crefname{subsection}{Sec.}{Secs.}

\title{Towards Continual Learning: Mitigating \\ Forgetting With Subspace Tuning}

\author{\textbf{Arnav Chavan}$^{\,1}$\hspace{0.2em}
\textbf{Kusha Sareen}$^{\,2,3}$\hspace{0.2em}
\textbf{Rishabh Tiwari}$^{\,1}$
\textbf{Abhranil Chandra}$^{\,5}$\hspace{0.2em}
\textbf{Johan Obando-Ceron}$^{\,2,4}$\hspace{0.2em}\\[0.2em]
\textbf{Kurt Keutzer}$^{\,1}$\hspace{0.2em}
\textbf{Scott Niekum}$^{\,5}$\hspace{0.2em}
\textbf{Inderjit S Dhillon}$^{\,6}$\hspace{0.2em}
\textbf{Devvrit Khatri}$^{\,7,8}$\\[0.8em]
$^{1}$UC Berkeley \quad
$^{2}$Mila \quad
$^{3}$McGill University \quad 
$^{4}$Universit\'e de Montr\'eal \quad
$^{5}$UMass Amherst \quad \\
$^{6}$Google \quad
$^{7}$UT Austin \quad
$^{8}$Mirendil}

\makeatletter
\def\addcontentsline#1#2#3{%
\addtocontents{#1}{\protect\contentsline{#2}{#3}{\thepage}{\@currentHref}{}}}
\makeatother
\begin{document}
\etocdepthtag.toc{mtoc}
\etocsettagdepth{mtoc}{none}
\etocsettagdepth{appendix}{none}

\newbox\dummybox
\setbox\dummybox\vbox{%
  \tableofcontents
}

\begin{tcolorbox}[abstractbox, width=\textwidth]
\maketitle
\begin{abstract}
Large language models (LLMs) acquire new skills through fine-tuning, but it can destroy capabilities they already have: catastrophic forgetting remains a central obstacle to continual learning in LLMs. Recent works
connect forgetting to how far, and along which directions, fine-tuning moves the weights.
On-policy methods such as reinforcement learning and on-policy distillation avoid updating the principal singular directions of weight matrices and thus forget less. In contrast, SFT concentrates updates along these directions and is susceptible to catastrophic forgetting. In this work, we introduce Subspace Tuning (SST), which restricts each weight update to the bottom singular subspace, eliminating changes along the principal directions. SST is optimizer-agnostic and composes well with fine-tuning recipes including SFT and self-distillation.
SST combines tight subspace protection, dense gradient estimation, and a simple refresh mechanism, enabling stronger learning than prior orthogonal fine-tuning methods.
With these mechanisms, protecting as much as 99\%
of the weight subspace substantially improves retention while preserving strong target-task performance.
This behavior strengthens with model and data size and nearly eliminates forgetting in the large-model regime. SST also keeps the model close to its base distribution, inducing $2$--$4\times$ less KL drift than the baseline recipes.
Overall, SST advances the learning--forgetting Pareto frontier, making it a practical default for fine-tuning pre-trained language models.
\end{abstract}
{\centering\small\authorFont\textbf{Correspondence}: \href{mailto:arnavchavan@berkeley.edu}{\texttt{arnavchavan@berkeley.edu}}, \href{mailto:devvrit.03@gmail.com}{\texttt{devvrit.03@gmail.com}}\par}
\end{tcolorbox}

\begin{figure}[h]
\centering
\includegraphics[width=0.99\textwidth]{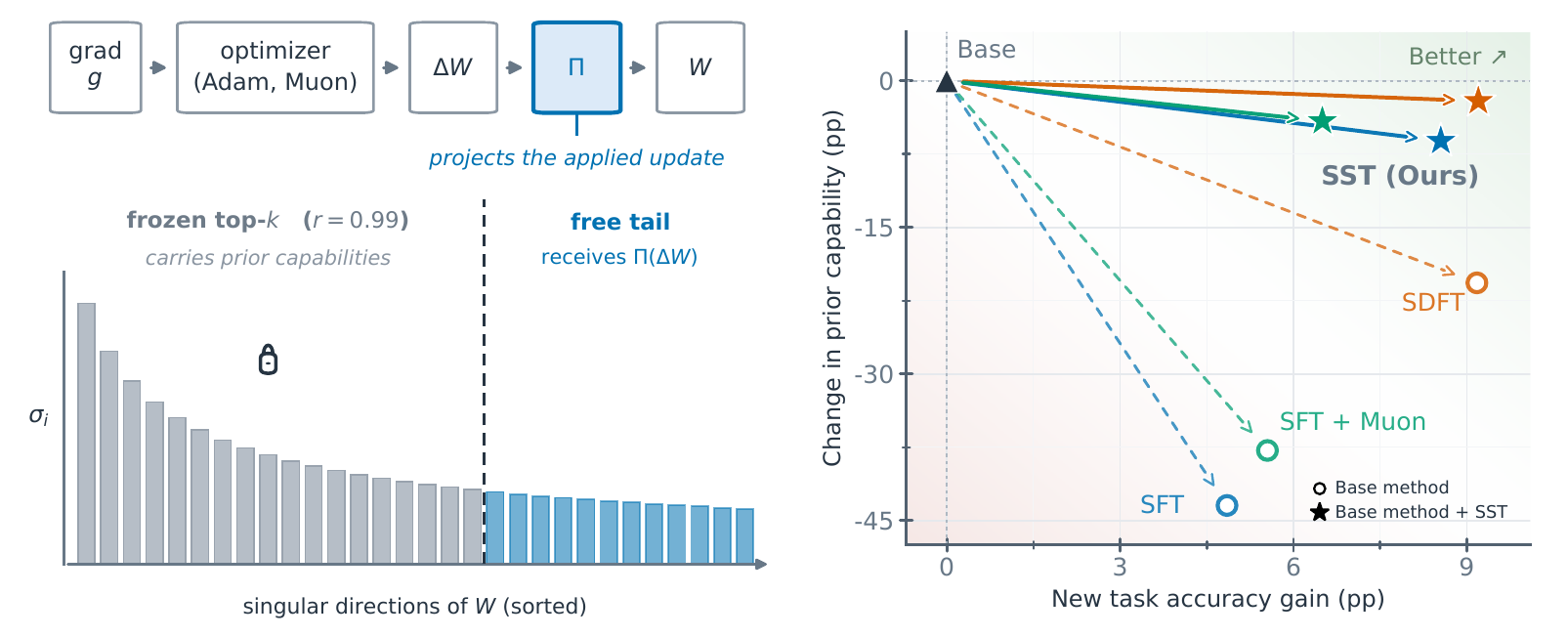}
\vspace{-1.6em}
\caption{\textbf{SST freezes the top of each weight matrix's spectrum, and improves learning--forgetting trade-off across fine-tuning recipes.} \textbf{Left:} the method. SST projects the update on low energy singular directions of
each weight matrix. \textbf{Right:} the result on Qwen3-8B, averaged over
four domains (Table~\ref{tab:qwen3-main}). Both axes show changes from the base model in percentage
points (pp); higher is better. Circles denote training without SST, stars denote training with SST,
and the black triangle marks the base model.}
\label{fig:pareto-avg}
\end{figure}

\vspace{-0.5em}
\section{Introduction}
\vspace{-0.2em}
\label{sec:introduction}

Large language models (LLMs) are useful across a rapidly growing range of applications, including assistants,
coding, search, science, and finance \citep{openai2023gpt4, handa2025economic,
chatterji2025chatgpt}. Many of these applications follow the same paradigm. A small number of strong base models are pre-trained once \citep{brown2020gpt3, touvron2023llama2, yang2025qwen3}, and
each use case then fine-tunes one of them on specialized data \citep{medpalm2,composer2technicalreport,devlin2019bert,
bommasani2021foundation, ouyang2022instructgpt}. Fine-tuning has thus become a standard way to specialize an LLM for a target task or domain.

This paradigm comes at the cost of catastrophic forgetting. As a model learns a new task,
capabilities acquired over trillions of training tokens, including instruction following, coding, reasoning, and general knowledge, can degrade severely~\citep{luo2025empiricalstudycatastrophicforgetting,lu2025onpolicydistillation,kotha2024}. This phenomenon is particularly difficult to address because the original pre-training data is typically unavailable~\citep{Kirkpatrick_2017}, while post-hoc repair may recover only part of what was lost~\citep{lu2025onpolicydistillation,shenfeld2026sdft} (\cref{sec:experiments}).
More broadly, forgetting stands in the way of continual learning. The next generation of foundation models should be able to improve over their deployed lifetime, adding new skills, domains, and corrections without sacrificing what they already do well. 

Recent works provide insights into the origin of this forgetting. The leading singular directions of pretrained weight matrices have been associated with important model functionality \citep{martin2021heavytailed,sharma2024laser}, while reinforcement learning has been found to fine-tune relatively small parameter subnetworks \citep{NEURIPS2025_bf235a1d}. More directly, RLVR updates concentrate outside the principal directions and induce relatively little spectral drift compared with SFT \citep{pathtakenrlvr}. On-policy RL also stays closer to the base policy and forgets less than SFT \citep{shenfeld2025razor}. These observations suggest a simple hypothesis. Preserving the principal singular subspaces of pretrained weights may preserve prior capabilities while leaving sufficient capacity elsewhere in weight space to learn new ones.
Existing methods induce this behavior indirectly. For example, SDFT \citep{shenfeld2026sdft} trains the model on its own rewrites of the training data, so that supervision stays close to the model's output distribution---approximately on-policy, and hence expected to forget less. This helps on non-reasoning instruct models, but we find substantial forgetting remains on longer reasoning traces (\cref{sec:experiments}).


We introduce \emph{Subspace Tuning} (SST), which restricts weight updates to a non-principal subspace of the weight spectrum.
Concretely, let $\Delta W$ be the optimizer's final update to a weight matrix $W$, and let $U_k$ and $V_k$ contain the top-$k$ left and right singular vectors of $W$. SST applies the projected update
\begin{equation}
\Pi(\Delta W) \;=\; (I - U_kU_k^\top)\,\Delta W\,(I - V_kV_k^\top)
\label{eq:proj-intro}
\end{equation}
in place of $\Delta W$. The two projectors remove from the update any component along the principal directions, on the output and input sides respectively, so every applied step stays in the non-principal subspace (\cref{sec:projection-formulation}).
Importantly, SST constrains the final optimizer update rather than the raw gradient. Stateful optimizers such as Adam rescale the gradient coordinate-wise before it reaches the weights, so a projected gradient does not produce a projected step and part of the update returns to the directions we mean to freeze (\cref{sec:method}). Because the projection acts on the applied update, SST is optimizer-agnostic and composes with any fine-tuning recipe, with no other change to training.

\begin{figure}[t]
\centering
\includegraphics[width=\textwidth]{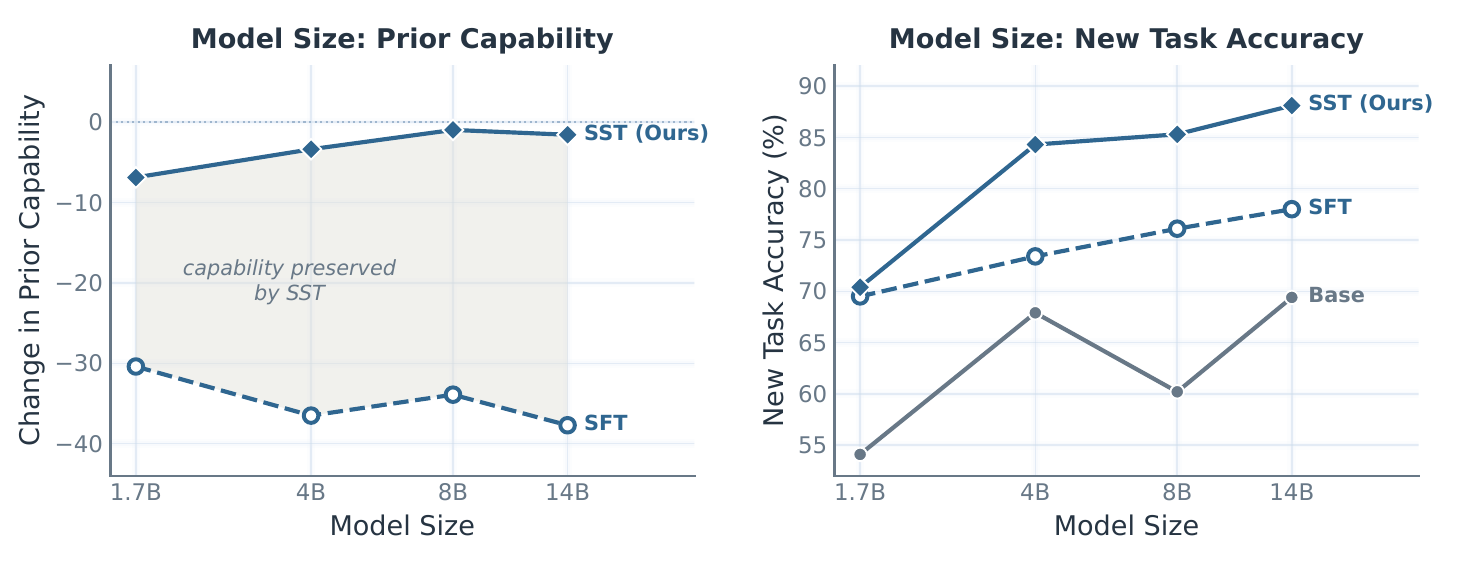}
\vspace{-0.7cm}
\caption{\textbf{Scaling model size.} Qwen3 models from $1.7$B to $14$B fine-tuned on Science for
five epochs. SFT forgets heavily at every scale, whereas forgetting under SST shrinks as models
grow and nearly vanishes from $8$B onward. SST also reaches higher new-task accuracy at every
size.}
\label{fig:scaling-size}
\end{figure}

\begin{figure}[t]
\centering
\includegraphics[width=\textwidth]{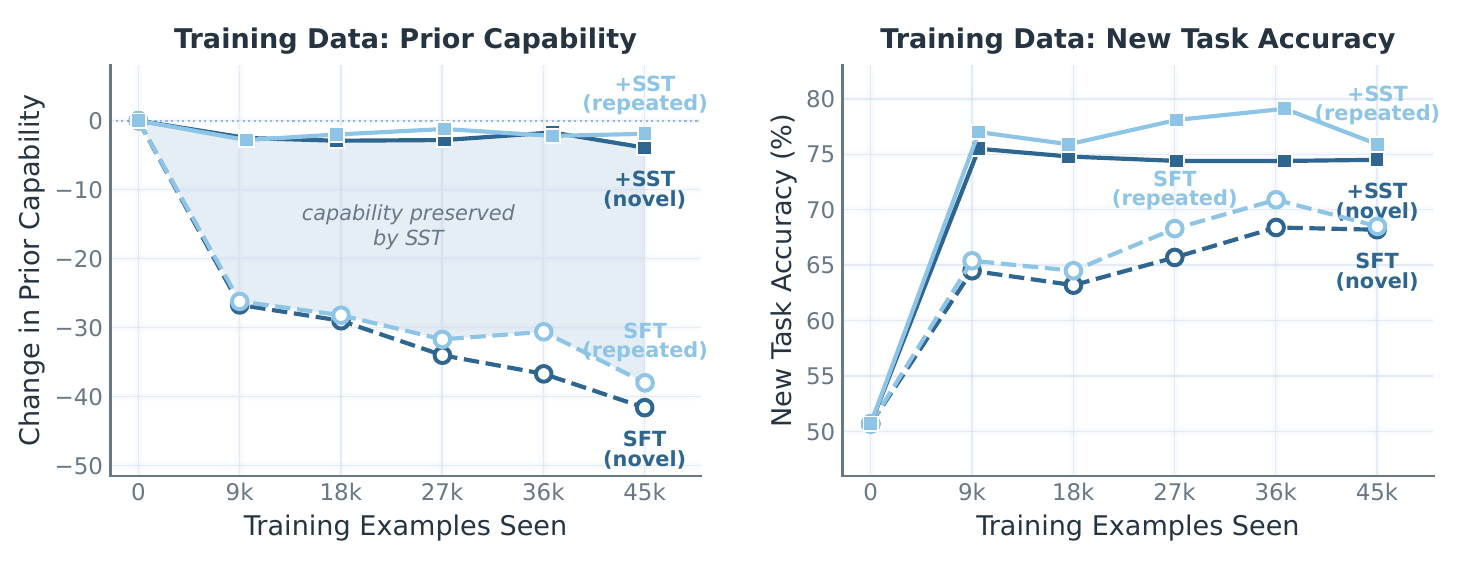}
\vspace{-0.7cm}
\caption{\textbf{Scaling training data.} Qwen3-8B fine-tuned on Science under a matched
$45$k-example budget: the \emph{novel} arms see a fresh stream once, the \emph{repeated} arms
cycle $9$k examples for five epochs. Under both exposures SFT drifts progressively farther from
the base model, while SST stays flat and learns the task better.}
\label{fig:scaling-data}
\end{figure}

We conduct extensive empirical analysis and establish the following properties of SST (\cref{sec:experiments}):
\begin{itemize}[leftmargin=*]
    \item \textbf{SST learns new tasks and retains prior capabilities.} Despite protecting the top $99\%$ of singular directions,
    SST matches the target-task accuracy of unconstrained fine-tuning, while retaining the prior capabilities that standard fine-tuning severely degrade (Figure~\ref{fig:pareto-avg}). This behavior holds across Science, Medicine, Pathology, Finance, and Tool Use, across SFT and SDFT, and with both Adam and Muon; Tables~\ref{tab:qwen3-main} and~\ref{tab:qwen25} provide the full per-domain results. We hypothesize that such constrained learning is possible because LLMs are heavily over-parameterized \citep{aghajanyan2021intrinsic,hu2022lora}, leaving enough capacity in the
    non-principal subspace
    to fit a new task. When a fixed complement becomes restrictive, a refresh step recomputes the SVD on the current weights, realigning the protected subspace so adaptation can continue (Section~\ref{sec:refreshing-basis}). \looseness=-1
    \item \textbf{SST scales well.} With more training data, SFT forgets progressively more while SST stays flat and still reaches higher target-task accuracy (Figure~\ref{fig:scaling-data}). With larger models, forgetting under SST shrinks even further, staying within $\sim 1\%$ of the base model from $8$B onward (Figure~\ref{fig:scaling-size}).
    \item \textbf{SST stays close to the base model.} It induces $2$-$4\times$ less KL divergence from the pretrained model than unconstrained recipes and preserves behavior that accuracy alone does not capture. A thinking model fine-tuned with SST keeps its natural reasoning length where SFT collapses it, and adding SST to SDFT reduces fabricated reference-solution attributions (Figure~\ref{fig:nothink-pair}).
    \item \textbf{SST makes sequential fine-tuning cumulative.} Over two- and three-stage fine-tuning chains of the above domains, each skill rises when it is trained and holds its level as later stages are learned, whereas under SFT each new stage degrades the dormant ones (Figure~\ref{fig:2stage-delta},\ref{fig:continual-mixed}).
\end{itemize}

Overall, this work shows that catastrophic forgetting is not an inevitable consequence of learning, but depends on where the update lands in weight space, and provides a simple, practical method for improving capability retention while acquiring new capabilities..

\vspace{-0.1cm}
\section{Subspace Tuning}
\vspace{-0.2cm}
\label{sec:method}
Across many complex systems, a small number of leading modes of a linear operator capture much of its dominant structure, while the remaining spectrum describes lower-energy variation \citep{berkooz1993pod, schmid2010dmd}. Deep neural network weight matrices exhibit similarly concentrated and heavy-tailed spectra \citep{martin2021heavytailed}, and interventions on their leading singular components can disproportionately affect model performance \citep{sharma2024laser}. These observations motivate preserving the principal singular directions of pretrained weights during adaptation.

We first formalize this constraint (\cref{sec:projection-formulation}), show that it must be enforced on the applied optimizer update rather than the gradient (\cref{sec:applying-projection}), and introduce a basis refresh that trades stricter protection for continued adaptability (\cref{sec:refreshing-basis}).

\subsection{Projection Formulation} \label{sec:projection-formulation}
Consider a single weight matrix $W \in \mathbb{R}^{d_\text{out}\times d_\text{in}}$, which maps an
input activation $x$ to an output $Wx$. Its SVD, $W=\sum_i \sigma_i\,u_i v_i^\top$, reads off the component of $x$ along the input direction $v_i$ (via $v_i^\top x$) and writes
it, scaled by $\sigma_i$, onto the output direction $u_i$. So the right singular vectors $v_i$ are the
\emph{input} directions the layer responds to, and the left singular vectors $u_i$ are the
\emph{output} directions it produces. The top-$k$ of these, $V_k=[v_1,\dots,v_k]$ and
$U_k=[u_1,\dots,u_k]$, are what we intend to protect. Let $P_U = U_kU_k^\top$ and $P_V = V_kV_k^\top$ be the orthogonal projectors onto the protected
output and input subspaces, and $Q_U = I - P_U$, $Q_V = I - P_V$ be their complements (the free
subspaces). Any output vector splits as protected-plus-free, $I = P_U + Q_U$, and likewise any input,
$I = P_V + Q_V$. Inserting these identities on the left and right of $\Delta W$ decomposes \emph{any}
update into four blocks, one per (input-side, output-side) combination:
\begin{equation}
\begin{aligned}
\Delta W ={}&
\underbrace{P_U\Delta W P_V}_{\text{(1) protected}\to\text{protected}}
+\underbrace{P_U\Delta W Q_V}_{\text{(2) free}\to\text{protected}}
+\underbrace{Q_U\Delta W P_V}_{\text{(3) protected}\to\text{free}}
+\underbrace{Q_U\Delta W Q_V}_{\text{(4) free}\to\text{free}} .
\end{aligned}
\label{eq:blocks}
\end{equation}

Freezing the top directions therefore means keeping only block~(4) and discarding (1)-(3) from the update $\Delta W$. Keeping only block~(4) is exactly the two-sided projection of Eq.~\ref{eq:proj-intro}, which we restate:
\begin{equation}
\Pi(\Delta W) \;=\; Q_U\,\Delta W\,Q_V \;=\; (I - U_kU_k^\top)\,\Delta W\,(I - V_kV_k^\top).
\label{eq:proj}
\end{equation}
In particular, the updated matrix $W_\text{ref} + \Pi(\Delta W)$ acts identically to $W_\text{ref}$
on every protected input direction and produces no new component along any protected output
direction: the protected singular triplets $(\sigma_i, u_i, v_i)$, $i \le k$, are preserved
exactly. Any of blocks (1)-(3) would break this, which is why the projection must be two-sided.
We fix $k = \lfloor r\cdot
\min(d_\text{out},d_\text{in})\rfloor$ with a rank fraction $r$ close to $1$ so only the low-energy
tail is free, and compute $U_k,V_k$ from $W_\text{ref}$, the weights at the start of training
(the anchor that refresh later updates, \cref{sec:refreshing-basis}).

\begin{figure}[t]
\centering
\includegraphics[width=\textwidth]{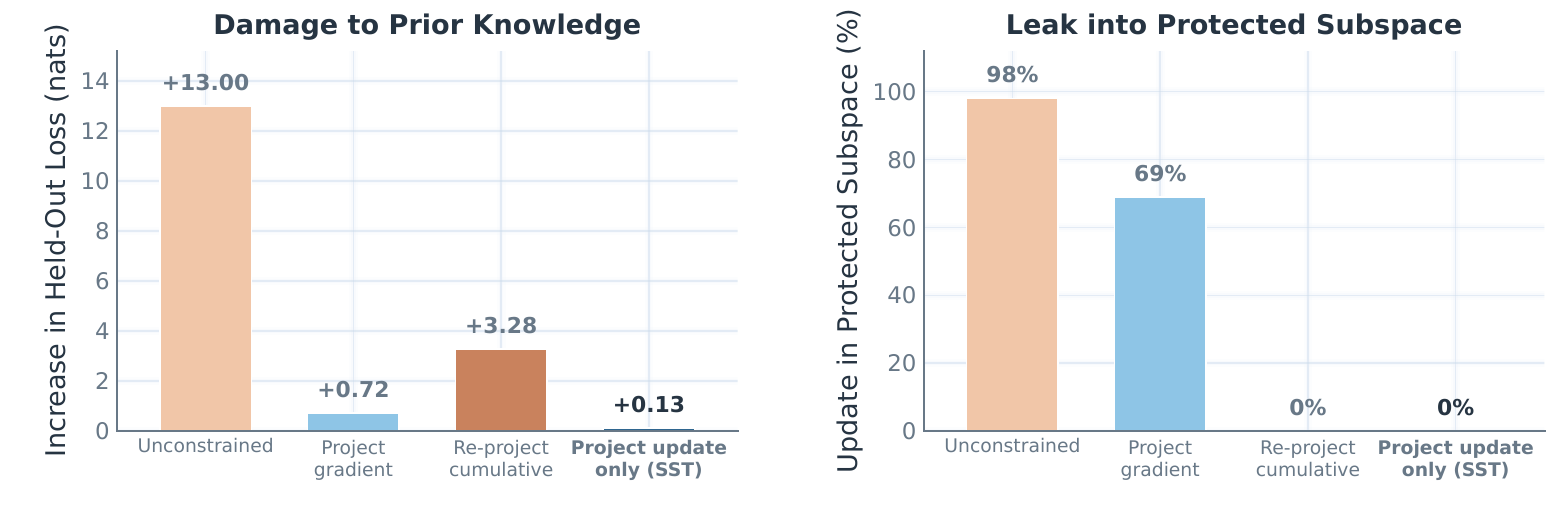}
\vspace{-0.8cm}
\caption{\small{\textbf{Controlled study on Qwen2.5-0.5B.}
Left: Increase in held-out NLL after fitting the toy task (lower is better). Right: Share of the
weight change entering protected directions (lower is better). Only update projection
achieves exact confinement while keeping the held-out NLL nearly unchanged.}}
\label{fig:toy-bars}
\vspace{-0.1cm}
\end{figure}

\subsection{Projecting the Optimizer Update}
\label{sec:applying-projection}

SST applies $\Pi$ to the optimizer's final update $\Delta W$. The projection could instead act on
the gradient before the optimizer, either alone or together with the applied update. These choices
are not equivalent. For plain SGD, whose step is a scalar multiple of the gradient, projecting the
gradient and projecting the update coincide. Under Adam and other stateful optimizers they differ.
Adam rescales each coordinate by its running moments, and this diagonal preconditioning does not
commute with $\Pi$, so a gradient confined to the free subspace can still produce an applied update
with a component in the protected subspace.

We compare three possible enforcement points (full equations in Appendix~\ref{app:toy}):
\begin{enumerate}[leftmargin=*,label=(\arabic*)]
\item \textbf{Project the gradient}: apply $\Pi$ to the raw gradient before Adam,
$g\leftarrow\Pi(g)$.
\item \textbf{Project the applied update}: let Adam update its moments from the full gradient, then
apply $\Pi$ to the step that Adam produces (this is what SST does).
\item \textbf{Re-project the cumulative change}: project the gradient $g$ and also project the
applied update $\Delta W$. This is equivalent to projecting the gradient and, after each step,
re-projecting the total change from the initial anchor, $W-W_\text{ref}$.
\end{enumerate}

To isolate this choice, we fine-tune Qwen2.5-0.5B for $60$ Adam steps on a fixed eight-example
next-token prediction task. We use $r{=}0.99$ and apply the constraint to the attention and MLP
linear maps. Because the task is deliberately easy, every method reaches a final training loss of
approximately $10^{-5}$. We can therefore compare their collateral effects using two measurements:
the increase in NLL on held-out generic sentences, and the fraction of the total weight change that
falls outside the allowed free-to-free block.

Our results are in Figure~\ref{fig:toy-bars}. Unconstrained fine-tuning increases the held-out
NLL by $13.0$, with $98\%$ of the weight change entering protected directions. Projecting the
gradient reduces the NLL increase to $0.72$, but Adam reintroduces substantial leakage: $69\%$ of
the final change still leaks into the protected directions. Re-projecting the cumulative change eliminates
leakage, but raises the held-out NLL by $3.28$. Projecting the applied update is the only scheme that
both eliminates leakage and keeps the probe nearly unchanged, with an NLL increase of just $0.13$.
We therefore design SST to project the final applied update. The full procedure is given in
Algorithm~\ref{alg:sst} in Appendix~\ref{app:toy}.



\vspace{-0.05cm}
\subsection{Refreshing the Protected Basis}
\label{sec:refreshing-basis}

The default variant computes $U_k,V_k$ once from the pretrained weights and keeps them fixed. During longer runs, the leading singular subspaces of the current weights can rotate away from this initial basis. SST can therefore re-anchor $W_\text{ref}$ to the current weights and recompute $U_k,V_k$ from it every $K$ optimizer steps, using a Gram-based eigendecomposition for efficiency (Appendix~\ref{sec:cost} discusses the minimal time overhead of SST compared to SFT).

A refresh defines the free complement relative to the current weights and opens new directions for adaptation. As the singular vectors rotate, a direction protected by the initial basis can acquire a component in a later free complement. Refresh can improve continued task learning, but it mildly weakens protection relative to the original model because SST now preserves a sequence of updated principal subspaces rather than one fixed pretrained subspace. We therefore treat $K$ as an adaptation--retention control and report no refresh, R100 ($K{=}100$), and R10 ($K{=}10$) through many of our experiments.

\vspace{-0.05cm}
\section{Experiments}
\label{sec:experiments}
We ask the central question:
\textit{When a model is fine-tuned on a new task, does confining the update to the non-principal subspace let it learn the task while retaining what it already knew?} 


\textbf{Models and tasks.}
We fine-tune Qwen3-8B \citep{yang2025qwen3} on data containing explicit \texttt{<think>}$\ldots$\texttt{</think>} traces in four domains: Science, Finance, Pathology, and Medicine. Science uses DeepSeek-R1-distilled traces from the \texttt{open-r1/Mixture-of-Thoughts} release \citep{openr1mot}. Finance uses the reasoning-annotated FinQA/ConvFinQA release \citep{finqa2021,convfinqa2022}. Pathology and Medicine use MedMCQA questions \citep{medmcqa2022} with traces generated by Claude Sonnet~4.6 \citep{anthropic2026sonnet46}.



\textbf{Evaluation.} \emph{In-distribution accuracy} (ID) measures how well the
new task was learned. \emph{Prior capability} measures forgetting: the average over held-out
general-purpose benchmarks not used for fine-tuning. We evaluate on a broad set of benchmarks before and after training, namely GPQA-Diamond \citep{gpqa}, MMLU-Pro \citep{mmlupro}, GSM8K \citep{gsm8k}, HumanEval \citep{humaneval}, and IFEval \citep{ifeval}.

\textbf{Baselines.} We compare against standard supervised fine-tuning (SFT) and self-distillation
fine-tuning \citep[SDFT,][]{shenfeld2026sdft}, an on-policy method designed to reduce forgetting.
We report SFT and SDFT alone, and each combined with SST at rank fraction
$r{=}0.99$ unless specified otherwise (Appendix~\ref{app:rank-finqa} sweeps $r$ and finds that
stronger protection improves both retention and in-domain accuracy), optionally with basis
refresh every $100$ (R100) or $10$ (R10) steps. We also include a
Muon-optimizer variant \citep{muon} and the post-hoc KL-repair baseline ReInvoke \citep{shenfeld2026sdft} where relevant.
Full training and evaluation details are in Appendix~\ref{app:hparams}.

\begin{figure}[h]
\centering
\includegraphics[width=0.85\textwidth]{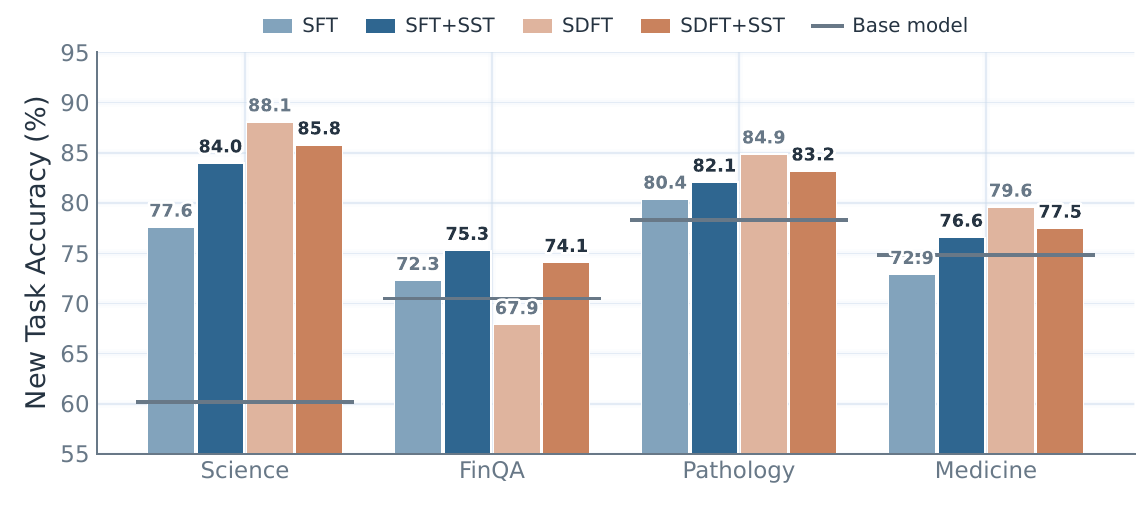}
\vspace{-0.7cm}
\caption{\textbf{Increase in in-domain accuracy} for
each recipe and its $+$SST$+$R10 version (Table~\ref{tab:qwen3-main}). Light bars are the recipe
alone, full-color bars the $+$SST version, and the gray rule in each group is the base
model. Refresh mechanism improves the in-domain accuracy matching SFT performance.}
\label{fig:in-domain}
\vspace{-0.3cm}
\end{figure}

\subsection{SST learns without forgetting}

\begin{figure}[h]
\centering
\includegraphics[width=\textwidth]{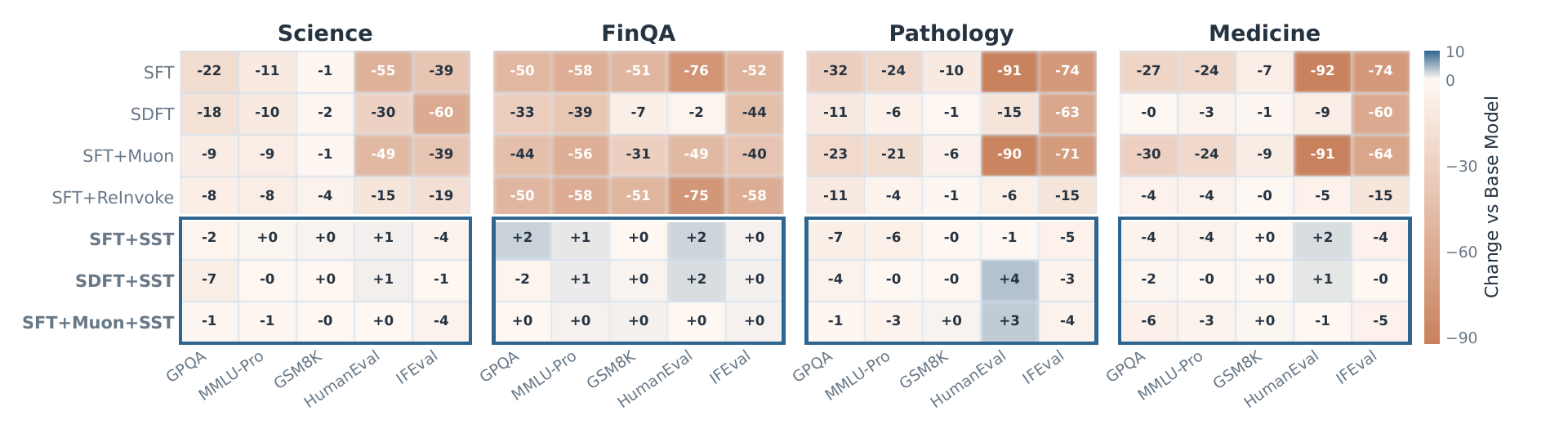}
\vspace{-0.7cm}
\caption{\textbf{Prior capability change compared to the base model} Unconstrained recipes incur broad capability losses. SST variants stay
much closer to the base model across benchmarks and domains.}
\label{fig:bench-heatmap}
\vspace{-0.1cm}
\end{figure}

We begin by fine-tuning a thinking model on external chain-of-thought data from a new domain. 

\emph{Unconstrained recipes collapse.} Figure~\ref{fig:bench-heatmap} shows broad losses in prior capability after both SFT and SDFT across the four domains. For example,
as the model learns the new task, it often stops producing compilable code or respecting requested output formats. Post-hoc KL repair recovers only part of this capability and reduces in-domain gains, leaving a learning--retention trade-off rather than improving both simultaneously.

\emph{SST retains prior capability.} Across domains, SST remains close to the base model while matching the in-domain performance of the corresponding unconstrained recipe. The benefit composes with SDFT and with the Muon optimizer, indicating it generalizes across
training recipes.
Figure~\ref{fig:pareto-avg} summarizes the improved learning--retention trade-off with R10 refresh. Figure~\ref{fig:bench-heatmap} gives the per-benchmark breakdown for the no-refresh variant, whose protection is strictest; switching to R10 costs only ${\sim}3-5$ additional points of prior capability on average (Table~\ref{tab:qwen3-main}). Figure~\ref{fig:in-domain} shows that with R10, retention does not come at the expense of target-task learning. More frequent refresh favors continued adaptation at a small retention cost, Appendix~\ref{app:qwen3-full} reports every variant.
A non-thinking instruct model, Qwen2.5-7B-Instruct, shows the same learning--retention pattern across three domains including Tool Use (Appendix~\ref{app:instruct}).

\vspace{-0.05cm}
\subsection{SST scales well}

\emph{SST reduces forgetting as parameters increase.} Figure~\ref{fig:scaling-size} shows the effect of scaling model size on the Science dataset, with each model trained for five epochs. Across model sizes, unconstrained SFT forgets substantially, losing $30$-$40$ points of prior capability. By contrast, SST preserves the base model better as scale grows. SST also remains competitive with or improves upon SFT on the new task, so the retention benefit is not obtained by suppressing learning. Training details are in Appendix~\ref{app:scaling}, and a separate two-epoch sweep through $32$B in Table~\ref{tab:scale} shows the same trend.

\emph{SST prevents monotonic decline in capability as the training data scales.} Figure~\ref{fig:scaling-data} shows the impact of scaling data on performance on prior and new tasks.
We compare two setting under a $45{,}000$-example budget. In the first, we have a novel stream of data and in the second we look at the multi-epoch setting: a $9$k example subset repeated for five epochs. In both settings, SFT moves progressively farther from the base model as training proceeds, while SST maintains prior capability and continues improving on the target task. SST therefore remains effective in both the single-epoch novel-data setting and the multi-epoch repeated-data setting. The corresponding table can be found in Appendix~\ref{app:scaling}.



\subsection{SST stays close to the base model}
\begin{figure}[t]
\centering
\includegraphics[width=0.9\textwidth]{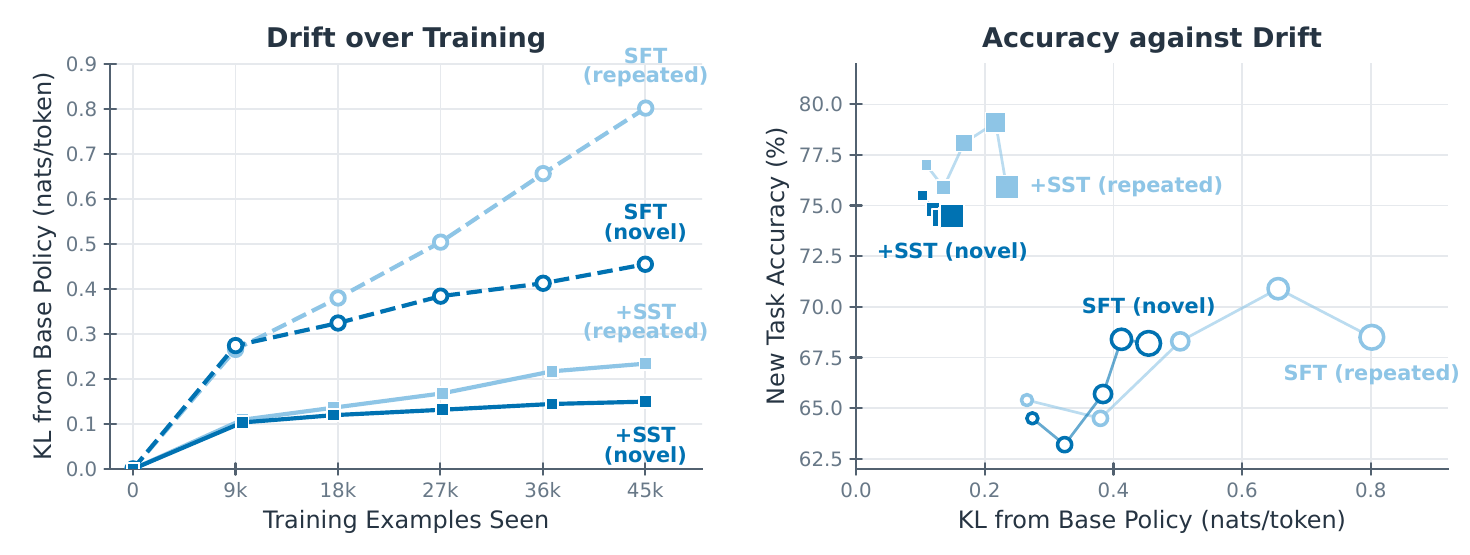}
\vspace{-0.2cm}
\caption{\textbf{Policy drift from the base model}, $\mathrm{KL}(\pi_{\text{base}}\,\|\,\pi_\theta)$ on
WikiText-2, a general-domain dataset containing no task data. Left: drift over training. Right: the same drift
against the in-domain accuracy, marker size growing with data seen.}
\label{fig:kl}
\vspace{-0.1cm}
\end{figure}

\vspace{-0.05cm}
\emph{SST keeps the model close to its base distribution.} Figure~\ref{fig:kl} measures divergence from the base model on WikiText-2 \citep{wikitext2}, a general-domain corpus. Across the training-data (Figure~\ref{fig:kl}) and model-scale studies (Figure~\ref{fig:kl-scale-bars}, Appendix~\ref{app:kl-scale}), SST remains substantially closer to the base distribution than the corresponding unconstrained model. The right panel of Figure~\ref{fig:kl} traces new-task accuracy against drift as training proceeds, and shows that SST improves accuracy with roughly $3\times$ less movement away from the base policy. Across model sizes and training recipes, the reduction is $2$-$4\times$ (Appendix~\ref{app:kl-scale}).

A natural alternative is to enforce closeness directly, by adding a reverse-KL penalty toward the base model during SFT. Appendix~\ref{app:kl-reg} (Table~\ref{tab:kl-reg}, Figure~\ref{fig:kl-reg}) shows that this does not recover SST's trade-off. Weak regularization fails to mitigate forgetting, and stronger regularization suppresses task learning without preserving prior capability.

\begin{figure}[t]
\centering
\includegraphics[width=0.86\textwidth]{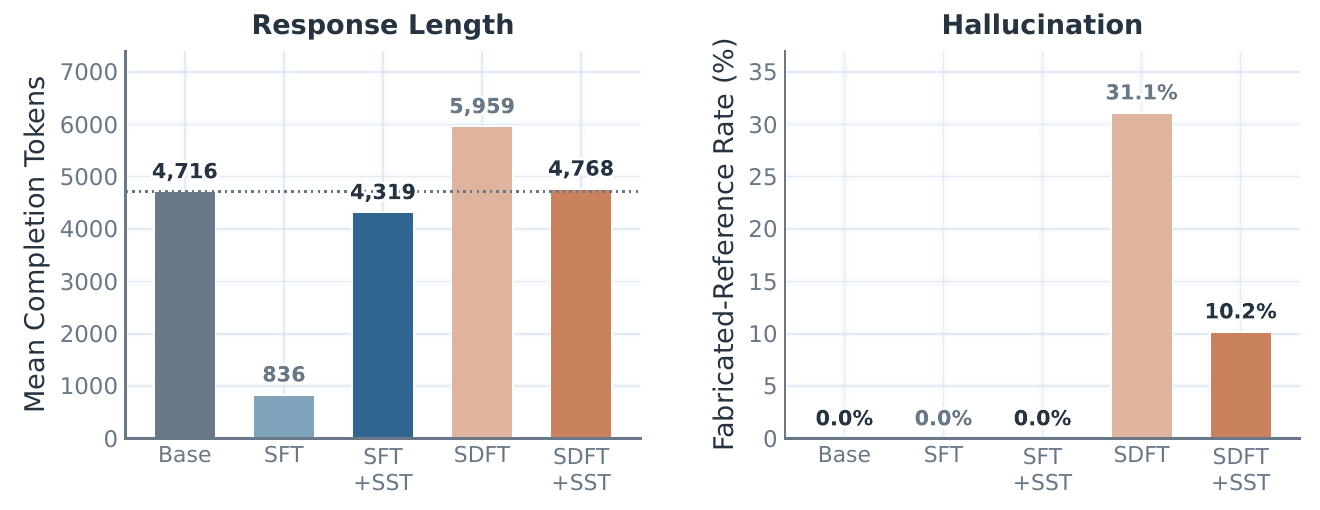}
\vspace{-0.5cm}
\caption{\textbf{Each recipe damages the base model in its own way, and SST pulls it back.}
Olmo-3-7B-Think fine-tuned on answer-only medical data ($+$R10 arms shown); the gray bar in each
panel is the base model. Left: Mean response length, with the dotted rule marking the base
model's response length. Right: Rate of fabricating a ``reference solution'' the prompt never
contained.}
\label{fig:nothink-pair}
\vspace{-0.1cm}
\end{figure}

\emph{SST reduces hallucinations induced by self-distillation.} 
Finally, we stress-test a difficult case for a reasoning model: finetuning with non-reasoning data. This setting explores a challenge where supervised finetuning can substantially degrade their reasoning capabilities.
We fine-tune Olmo-3-7B-Think \citep{olmo3} on answer-only medical supervision from HuatuoGPT-o1 \citep{huatuogpto1} and summarize the results in Figure~\ref{fig:nothink-pair}. SFT improves answer accuracy but sharply shortens the model's reasoning, while SST keeps its response length close to the base model. SDFT maintains longer reasoning but can overshoot and fabricate references to a solution that was never provided. Adding SST's subspace constraint substantially reduces this behavior. Together, these results show that staying close to the base model preserves qualitative properties not captured by task accuracy alone. The plotted fabrication rate is a conservative phrase-based lower bound. Appendix~\ref{app:hparams} gives its exact definition and the grading protocol.

\vspace{-0.05cm}
\subsection{SST preserves gains across successive fine-tuning stages}
\vspace{-0.05cm}

A stronger test is whether SST lets a model acquire several skills in sequence without each stage
erasing the last. We fine-tune Qwen3-8B on two consecutive tasks ($A\rightarrow B$), tracking retention on task $A$ and learning on task $B$. We also evaluate three-stage chains ($A\rightarrow B\rightarrow C$) and track each domain throughout.

\begin{figure}[t]
\centering
\includegraphics[width=\textwidth]{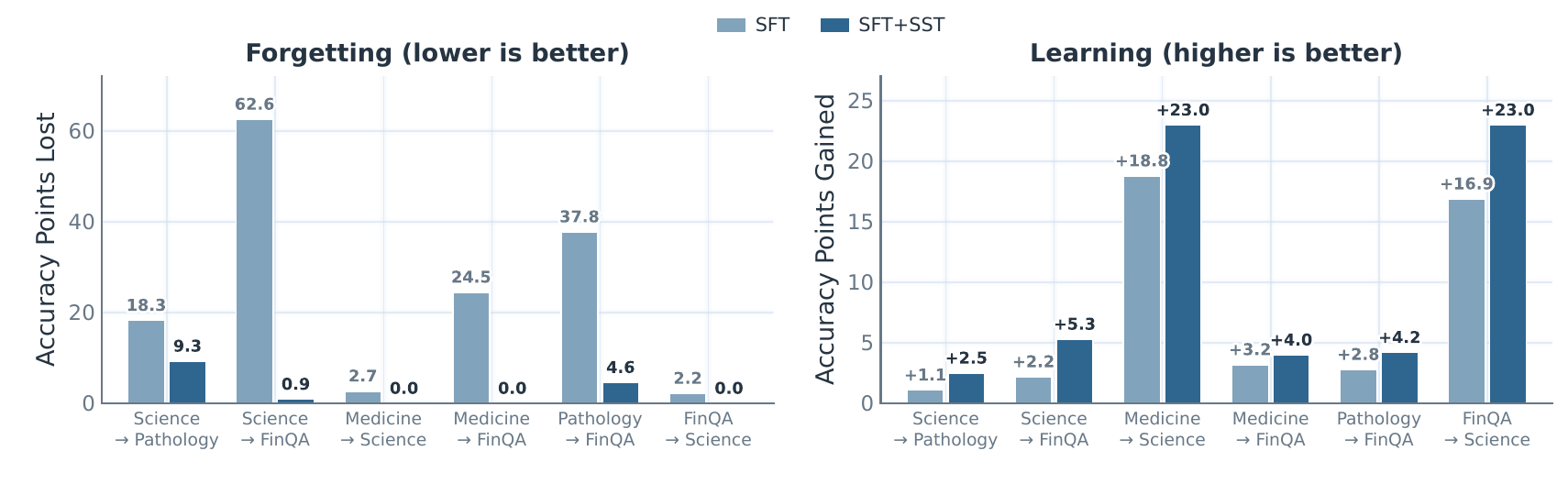}
\vspace{-0.8cm}
\caption{\textbf{Two-stage continual fine-tuning ($A\rightarrow B$) on Qwen3-8B.} Left: accuracy lost on domain $A$ relative to its stage-1 level while stage 2 trains on $B$. Right: accuracy gained on $B$ over the base model. SST arms use R10 refresh. SST improves \emph{both} quantities in every tested chain.}
\label{fig:2stage-delta}
\vspace{-0.05cm}
\end{figure}

\emph{SST learns more and forgets less.} Figure~\ref{fig:2stage-delta} shows multiple $A\rightarrow B$ sequences across domain pairs. We use R10 refresh in these runs. Across all the tested combinations, SST outperforms SFT in both retention and new-task learning. Appendix~\ref{app:continual-sdft-two-stage} adds SDFT to three of these chains. SDFT learns each target well but its retention is chain-dependent, for example losing $20$ points on FinQA $\rightarrow$ Science where SST+R10 loses none.

\begin{figure}[t]
\centering
\includegraphics[width=\textwidth]{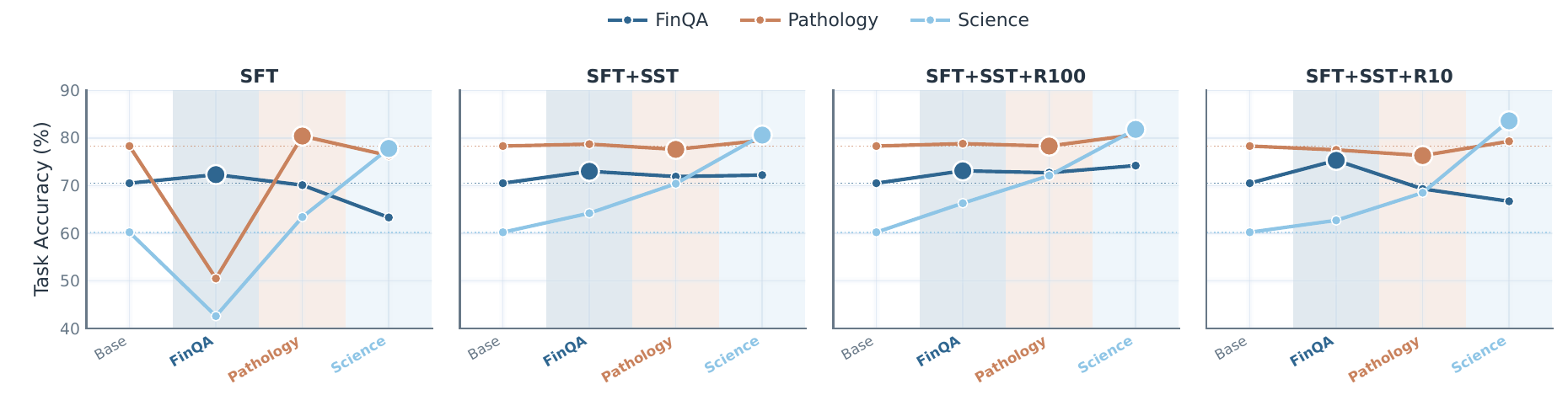}
\vspace{-0.7cm}
\caption{\textbf{Three-stage continual fine-tuning on Qwen3-8B.} The SFT family is trained on
FinQA $\rightarrow$ Pathology $\rightarrow$ Science. Lines track task accuracy on all three domains,
dotted lines mark each domain's base-model accuracy, each stage's background is tinted with the
domain trained at that stage, and the enlarged marker is the domain just trained. A line should
rise in its own stage and hold its level afterward. Under SFT the dormant domains drop sharply and
recover only when trained, while the SST variants hold all three domains much closer to their
trained levels.
}
\label{fig:continual-mixed}
\end{figure}

\emph{SST accumulates task knowledge.} In the FinQA $\rightarrow$ Pathology $\rightarrow$ Science chain of Figure~\ref{fig:continual-mixed}, each domain improves when it is trained and remains near that level as later stages are learned, so the final model retains all of the acquired skills. Under plain SFT the same chain is not cumulative. Accuracy on the untrained Pathology and Science drops by $28$ and $18$ points while FinQA trains, and FinQA itself gives back $9$ points over the later stages. The complementary Medicine $\rightarrow$ Pathology $\rightarrow$ Science chain in Appendix~\ref{app:continual-sdft} shows the same pattern for the SDFT family. The refresh variants expose a useful trade-off in this sequential setting. More frequent refresh improves accuracy on the current task at a minor cost to retention on previously learned tasks.

As a complementary composition test, Appendix~\ref{app:merge} shows that independently trained SST experts can be merged with far less interference than SFT experts. Figure~\ref{fig:merge} reports the exact task-level changes relative to each method's own experts.

\vspace{-0.1cm}
\section{Related Work}
\label{sec:relwork}
\vspace{-0.1cm}

Classical continual learning restricts where updates go using stored data or gradients from
earlier tasks \citep{Kirkpatrick_2017, lopezpaz2017gem, chaudhry2019agem, farajtabar2020ogd,
saha2021gpm, wang2021adamnscl}. These constraints require data that is unavailable at LLM scale
and task boundaries that deployed fine-tuning does not have, whereas SST needs neither.
A complementary line reduces forgetting through the data instead of the update.
Self-distillation rewrites supervision into the model's own distribution \citep{shenfeld2026sdft},
and on-policy RL forgets less because its updates stay close to the base policy
\citep{shenfeld2025razor, pathtakenrlvr}. Those methods induce a favorable update geometry through
on-policy data. SST, by contrast, constrains weight update, and as a result can compose with on-policy methods (Table~\ref{tab:qwen3-main}). We defer discussion of many other related works to Appendix~\ref{sec:related}.

\subsection{Comparison with subspace-protecting methods}
\label{sec:subspace-baselines}

Two recent methods share our starting point of protecting the top singular directions of
pretrained weights while adapting elsewhere. Both differ from SST in the same structural way.
They rewrite the update as a product of trained factors and constrain the factors, whereas SST
trains the weights directly and constrains the one quantity that reaches them, the applied
update.

Sculpting \citep{nayak2025sculpting} freezes the top singular vectors of each weight matrix and
projects the gradient before every AdamW step. Its projection removes only block (1) of
Eq.~\ref{eq:blocks}, the component that reads from and writes onto protected directions at once.
The cross blocks (2) and (3), which touch a protected direction on one side, are never removed.
The projection moreover acts on the gradient, rather than the update as suggested by our
controlled study. Even removing all three blocks from the gradient leaves $69\%$ of the applied
update in protected directions (Figure~\ref{fig:toy-bars}).

OPLoRA \citep{xiong2025oplora} freezes the weights and trains a LoRA adapter, the product $BA$ of
two low-rank trainable matrices, sandwiched between fixed projectors, $\Delta W = Q_U\, BA\, Q_V$
in our notation. This is the projection $\Pi$ of
Eq.~\ref{eq:proj} applied to a low-rank product rather than to the optimizer's step, and it
confines the update fully to the free subspace. The amount
protected is set by the projector rank $k$, and their experiments protect only the top
$16$-$128$ directions while adapting at low rank in everything that remains, nearly the inverted
regime from SST, which protects $99\%$ of the spectrum and adapts densely in the tail. By the
chain rule through the fixed projectors, the gradient components along protected directions never
reach the trainable factors, and the applied update is projected again by construction. This
corresponds to scheme (3) in \cref{sec:applying-projection}, which our controlled study found
mis-scales the step (Figure~\ref{fig:toy-bars}), with Adam's step sizes moreover set in factor
space rather than for the weight update.
\looseness=-1

Another difference is that neither method considers refreshing the protected subspace.
SST introduces refresh as part of the method, re-anchoring the
basis every $K$ steps, and studies it as an explicit retention-adaptation control - swept from no
refresh to R10 throughout our experiments (\cref{sec:refreshing-basis}).

We compare both methods on the four-domain setup under their official settings
(Table~\ref{tab:baselines-4dom}). Both retain more capability than SFT but lose $18$-$26$ points
of prior capability on average where no-refresh SST loses about one, at comparable in-domain
accuracy. Giving OPLoRA our protection level ($k{=}0.99d$) recovers much of the gap, yet it still
trails SST on three of the four domains (Table~\ref{tab:qwen3-main}).
Under data scaling across Medicine, Pathology, and FinQA with official settings, both OPLoRA and
Sculpting forget substantially more as training continues, while SST stays much closer to the
base model (Appendix~\ref{app:baselines}, Figure~\ref{fig:finqa-datascale}). We therefore keep SFT and SDFT, the dominant fine-tuning
recipes in practice, as the primary baselines throughout the paper, while comparing against
Sculpting and OPLoRA in this main setting and on data scaling.
\looseness=-1

\vspace{-0.2cm}
\section{Conclusion}
\label{sec:conclusion}

Catastrophic forgetting is usually treated as the price of learning new tasks. Our work studies it instead as a
property of \emph{where} the update lands, directly controls it, and mitigates forgetting. Confining the applied
optimizer step to the complement of each weight matrix's top singular subspace (Eq.~\ref{eq:proj}) turns
recipes that destroy prior capability into ones that retain it. The same
operator works across recipes (SFT, SDFT, and Muon), from $1.7$B to $32$B parameters, and across data scale, where SFT degrades monotonically and SST stays flat. It also turns sequential fine-tuning from destructive into cumulative and keeps the model $2$--$4\times$ closer to its base
distribution in KL, thus preserving behavior that accuracy alone does not capture, such as a thinking
model's deliberation length and self-distillation's fabricated reference
solutions.

\textbf{Limitations and Future Work.}
SST changes the feasible update geometry, its optimal
learning rate need not equal that of unconstrained SFT. Establishing fully
tuned learning--retention frontiers remains important future work. Scaling SST to models with hundreds of billions of
parameters and to longer task sequences is an important next test. Additionally, using information about the update for each batch to dynamically update the protection rather than refreshing at fixed intervals is a next step to make the algorithm more efficient.


\bibliography{iclr2027_conference}
\bibliographystyle{plainnat}

\newpage
\appendix
\onecolumn

\etocdepthtag.toc{appendix}
\etocsettocstyle{\section*{Appendix Contents}}{} 
\setcounter{tocdepth}{3}
\etocsettagdepth{mtoc}{none}           
\etocsettagdepth{appendix}{subsection}  
\tableofcontents                        

\clearpage

\section{Extended Related Work}
\label{sec:related}

Classical continual-learning approaches often restrict the update direction: regularization penalizes movement along Fisher-important directions
\citep{Kirkpatrick_2017}, while projected-gradient methods enforce constraints by replacing gradient $g$ with a projection $\Pi(g)$ built from stored data, past-task gradients
\citep{lopezpaz2017gem, chaudhry2019agem, farajtabar2020ogd}, activation subspaces
\citep{saha2021gpm}, the feature-covariance null space \citep{wang2021adamnscl}, or earlier tasks'
adapters \citep{wang2023olora}. Continual learning for LLMs differs from these settings in several ways. These methods estimate subspaces from data, but reconstructing a distribution representative of an LLM's pretraining corpus is infeasible at scale. Truly continual deployment may also lack explicit task boundaries. SST instead preserves a subspace derived from the weight spectrum (Eq.~\ref{eq:proj}), requiring neither stored examples nor task boundaries.

Several parameter-efficient fine-tuning methods also construct adapters from singular directions. PiSSA \citep{meng2024pissa} initializes and trains LoRA factors from the principal singular components, prioritizing high-energy directions rather than protecting them. MiLoRA \citep{wang2025milora} is closer in motivation: it retains the principal components and adapts minor singular components, but represents the update through a fixed low-rank factorization. CorDA \citep{yang2024corda} first forms a context-oriented decomposition using input-activation covariance and initializes low-rank adapters from task- or knowledge-oriented components. In all three methods, the spectral choice is encoded in a parameter-efficient factorization derived at initialization. SST instead projects the dense update actually proposed by the optimizer into the two-sided non-principal complement. It therefore constrains the realized weight change after stateful optimizer preconditioning, does not rely on a trainable adapter factorization or task-specific covariance estimate, and can refresh its protected basis during training.

A complementary line leaves the optimizer
alone and reduces the distribution shift driving forgetting: self-distillation rewrites the supervision
into the model's own output distribution \citep{shenfeld2026sdft}, and online RL forgets less because
its updates stay KL-close to the base policy \citep{shenfeld2025razor, tiwari2026learningfastslowllms}, both consistent with
forgetting tracking distributional movement \citep{kotha2024,
luo2025empiricalstudycatastrophicforgetting}. These methods act on the data, whereas SST acts on the update, and the two
compose (Table~\ref{tab:qwen3-main}).

\section{Additional Experiments \& Details}
\label{sec:appendix}
This appendix contains additional results supporting the main text. In particular, we report extra results on model merging independently fine-tuned experts, the Qwen2.5 instruct-model, complete model- and data-scaling results, full Qwen3 tables, policy drift across model
sizes, the toy study and algorithm specification, comparisons with Sculpting and OPLoRA, 
reverse-KL and protected-rank controls, and full training and evaluation details.

\subsection{Model Merging}
\label{app:merge}

If a constrained update stays in the low-energy complement of the weight spectrum, then two
updates trained independently should interfere less when added back together. We test this with model merging experiments. In particular, we fine-tune a separate single-domain expert for Science, Medicine, and FinQA, then form
$W_{\mathrm{merged}}=W_{\mathrm{base}}+\sum_d (W_d-W_{\mathrm{base}})$ over all the $252$ weight
matrices. We copy every other parameter from the base model. No merge coefficients are tuned and no
data is used to fit the combination. Each merged model is then scored in-domain against \emph{its own}
single-domain expert under identical decode settings. We report the interference caused by
merging rather than the raw accuracy gap, which already favors SST.

\begin{figure}[htbp]
\centering
\includegraphics[width=0.55\textwidth]{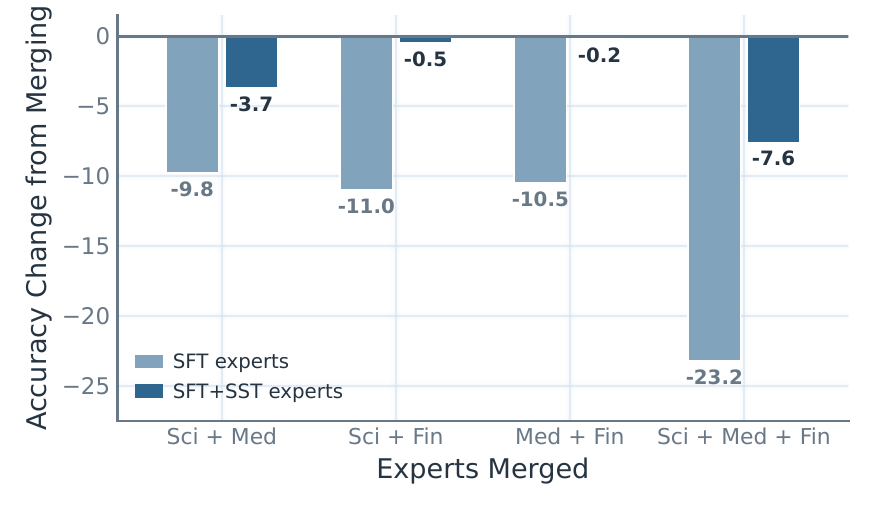}
\caption{Interference from task-arithmetic merging. Each bar is the mean change in task accuracy
relative to the same method's own single-domain experts.}
\label{fig:merge}
\end{figure}

\emph{SST experts compose, SFT experts collide.} Figure~\ref{fig:merge} shows that independently trained SFT experts interfere strongly when their task vectors are combined, and the interference compounds as more experts are merged. SST experts compose much more cleanly: pairwise merges remain close to their corresponding single-domain experts, and the three-expert merge degrades substantially more slowly. Each SST expert modifies only the low-energy subspace, so adding their updates leaves the protected principal directions unchanged. Unconstrained experts can each alter these directions, causing their changes to accumulate during merging.

\subsection{Two-stage continual learning with SDFT}
\label{app:continual-sdft-two-stage}

\begin{figure}[htbp]
\centering
\includegraphics[width=\textwidth]{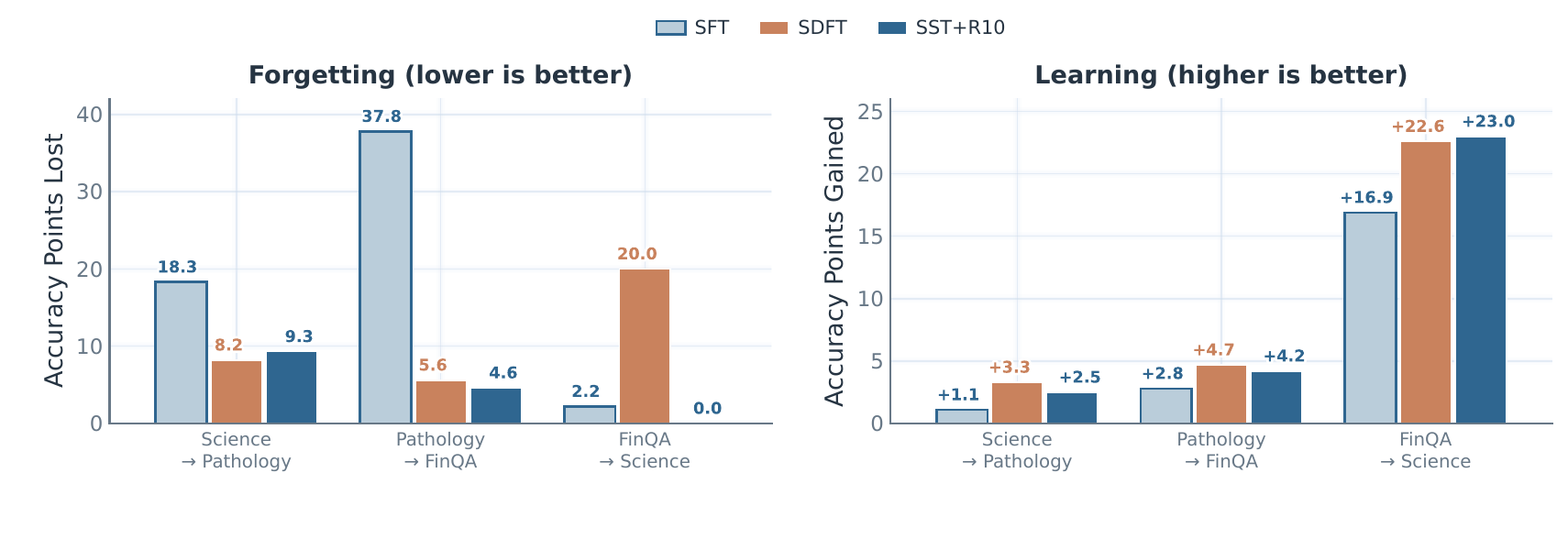}
\caption{\textbf{Two-stage continual fine-tuning with SDFT ($A\rightarrow B$) on Qwen3-8B.}
Left: accuracy lost on domain $A$ relative to its stage-1 level after training on $B$.
Right: accuracy gained on $B$ over the base model. All arms use the same task chains and
stage budgets; SST+R10 denotes SFT with SST and R10 refresh.}
\label{fig:continual-2stage-sdft}
\end{figure}

Figure~\ref{fig:continual-2stage-sdft} adds plain SDFT to the same two-stage setup as
Figure~\ref{fig:2stage-delta}. SDFT improves target learning over SFT in all three chains,
but its retention is chain-dependent. It reduces forgetting on Science $\rightarrow$ Pathology
and Pathology $\rightarrow$ FinQA, yet loses $20.0$ points on FinQA $\rightarrow$ Science.
SST+R10 gives the most consistent retention--learning trade-off.

For Pathology $\rightarrow$ FinQA, we regrade the saved SDFT generations by prioritizing the final
boxed numeric answer before the existing \texttt{Answer:} fallback. This corrects an
evaluator-format mismatch, giving $75.2$ target accuracy ($+4.7$ over base); the source-task score,
and therefore the $5.6$ points of forgetting, is unchanged.

\subsection{Three-stage continual learning with SDFT}
\label{app:continual-sdft}

\begin{figure}[htbp]
\centering
\includegraphics[width=\textwidth]{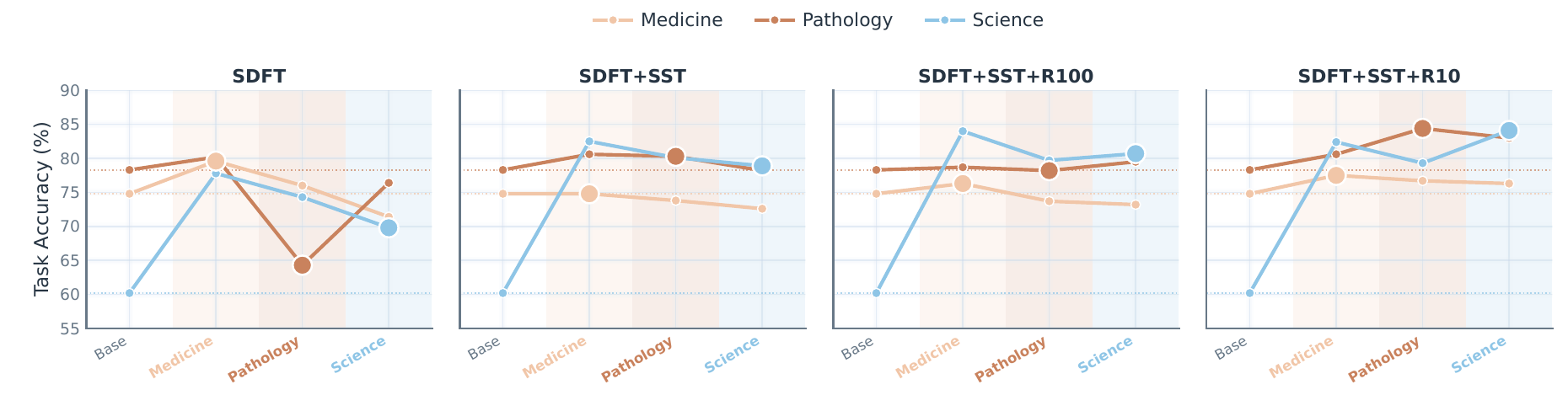}
\caption{\textbf{Three-stage SDFT chain on Qwen3-8B.} The SDFT family is trained on
Medicine $\rightarrow$ Pathology $\rightarrow$ Science. Lines track all three domains, dotted
lines mark base-model accuracy, and the background tint identifies the domain trained at each
stage.}
\label{fig:continual-sdft}
\end{figure}

Figure~\ref{fig:continual-sdft} shows the Medicine $\rightarrow$ Pathology
$\rightarrow$ Science continual learning chain discussed in the main text on SDFT. Across the three stages, plain SDFT consistently degrades performance on
Medicine and Science. SST stabilizes these trajectories, especially the R10 variant accumulates learning across all three domains.

\subsection{Instruct tuning without forgetting}
\label{app:instruct}

We also study a non-thinking instruction model with shorter sequences, where forgetting is less severe. Table~\ref{tab:qwen25} reports Qwen2.5-7B-Instruct \citep{qwen25} on ToolAlpaca \citep{toolalpaca}, HuatuoGPT-o1 \citep{huatuogpto1}, and SciKnowEval \citep{sciknoweval}, matching the SDFT setup \citep{shenfeld2026sdft}. All cells use rank fraction $r{=}0.99$ except Science, which uses $r{=}0.90$. On this model and domain, the tail left free at $r{=}0.99$ was too small to fit the task. This is the only setting in the paper that required a looser protected rank.

\emph{SFT learns the task but forgets.} Table~\ref{tab:qwen25} shows that standard fine-tuning improves the target task but reduces prior capability, with the clearest losses in instruction following and knowledge. SDFT moderates this effect in some settings but is not consistently protective across domains.

\emph{SST holds capability at matched fit.} Adding SST recovers much of the lost prior capability on Tool Use and Medical while remaining competitive on in-domain accuracy. Science is a narrower trade-off: SST variants can improve task learning, but no single variant dominates both learning and retention. Table~\ref{tab:qwen25} gives the exact per-domain and per-benchmark values. The capability average covers HellaSwag \citep{hellaswag}, TruthfulQA-MC1 \citep{truthfulqa}, MMLU \citep{mmlu}, IFEval \citep{ifeval}, and WinoGrande \citep{winogrande}.

\begin{table}[htbp]
\centering
\caption{Qwen2.5-7B-Instruct. ID $=$ in-distribution accuracy; LM average over
HellaSwag, TruthfulQA-MC1, MMLU, IFEval, and WinoGrande; $\Delta$ is relative to the
pretrained baseline ($68.5$). Bold entries mark column maxima among non-base recipes within each domain.}
\label{tab:qwen25}
\footnotesize
\setlength{\tabcolsep}{2.6pt}
\resizebox{\textwidth}{!}{%
\begin{tabular}{llccccccc}
\toprule
Domain & Method & ID & HellaSwag & TQA-mc1 & MMLU & IFEval & Winogr. & LM avg ($\Delta$) \\
\midrule\midrule
\multirow{9}{*}{ToolUse}
 & Baseline             & ---  & 80.5 & 47.6 & 71.8 & 71.5 & 71.0 & 68.5 ($+0.0$) \\
\cmidrule{2-9}
 & SFT                  & 70.1 & \textbf{74.3} & 39.5 & 57.7 & 45.7 & \textbf{68.7} & 57.2 ($-11.3$) \\
 & SFT+SST      & 70.1 & 65.5 & \textbf{47.6} & 68.4 & 70.4 & 62.6 & 62.9 ($-5.6$) \\
 & SFT+SST+R100 & 69.1 & 65.3 & 47.2 & 68.3 & 69.5 & 63.0 & 62.7 ($-5.8$) \\
 & SFT+SST+R10  & 71.1 & 67.6 & 47.4 & 68.5 & 69.7 & 63.3 & 63.3 ($-5.2$) \\
\cmidrule{2-9}
 & SDFT                 & 70.1 & 70.8 & 44.2 & \textbf{68.9} & 65.8 & 67.7 & 63.5 ($-5.0$) \\
 & SDFT+SST     & 66.0 & 66.7 & 43.5 & 68.5 & 71.3 & 64.3 & 62.9 ($-5.6$) \\
 & SDFT+SST+R100& 64.9 & 67.0 & 44.6 & 68.8 & 71.7 & 63.8 & 63.2 ($-5.3$) \\
 & SDFT+SST+R10 & \textbf{75.3} & 68.2 & 44.7 & 68.8 & \textbf{74.5} & 63.5 & \textbf{63.9} ($-4.6$) \\
\midrule\midrule
\multirow{9}{*}{Medical}
 & Baseline             & ---  & 80.5 & 47.6 & 71.8 & 71.5 & 71.0 & 68.5 ($+0.0$) \\
\cmidrule{2-9}
 & SFT                  & 20.5 & \textbf{72.3} & 36.5 & 68.5 & 61.9 & 67.6 & 61.4 ($-7.1$) \\
 & SFT+SST      & 19.5 & 70.8 & 41.4 & 69.6 & 68.8 & 68.3 & \textbf{63.8} ($-4.7$) \\
 & SFT+SST+R100 & 19.0 & 71.1 & 41.2 & 69.3 & 68.2 & 67.6 & 63.5 ($-5.0$) \\
 & SFT+SST+R10  & 19.2 & 70.5 & 41.7 & \textbf{70.1} & 68.9 & 67.0 & 63.7 ($-4.8$) \\
\cmidrule{2-9}
 & SDFT                 & \textbf{29.7} & 67.1 & 39.7 & 55.7 & 58.8 & \textbf{68.4} & 57.9 ($-10.6$) \\
 & SDFT+SST     & 26.9 & 68.0 & 40.5 & 69.0 & \textbf{73.0} & 68.0 & 63.7 ($-4.8$) \\
 & SDFT+SST+R100& 29.6 & 68.1 & 41.7 & 69.2 & 72.1 & 68.1 & \textbf{63.8} ($-4.7$) \\
 & SDFT+SST+R10 & 28.0 & 67.3 & \textbf{41.9} & 68.3 & \textbf{73.0} & 67.2 & 63.5 ($-5.0$) \\
\midrule\midrule
\multirow{9}{*}{Science}
 & Baseline             & ---  & 80.5 & 47.6 & 71.8 & 71.5 & 71.0 & 68.5 ($+0.0$) \\
\cmidrule{2-9}
 & SFT                  & 68.6 & \textbf{73.2} & 43.0 & 69.2 & 61.4 & 68.2 & \textbf{63.0} ($-5.5$) \\
 & SFT+SST      & 67.7 & 71.3 & 42.5 & \textbf{70.6} & 63.8 & 66.1 & 62.8 ($-5.7$) \\
 & SFT+SST+R100 & 67.5 & 71.4 & 42.4 & 69.7 & 63.6 & 66.8 & 62.8 ($-5.7$) \\
 & SFT+SST+R10  & 69.8 & \textbf{73.2} & 42.2 & 68.2 & 55.3 & 67.2 & 61.2 ($-7.3$) \\
\cmidrule{2-9}
 & SDFT                 & 69.0 & 70.1 & \textbf{43.7} & 67.2 & 61.4 & 66.5 & 61.8 ($-6.7$) \\
 & SDFT+SST     & 67.1 & 68.9 & 43.5 & 67.4 & 62.5 & 66.4 & 61.7 ($-6.8$) \\
 & SDFT+SST+R100& \textbf{70.8} & 69.3 & 42.8 & 66.1 & \textbf{64.1} & 68.0 & 62.1 ($-6.4$) \\
 & SDFT+SST+R10 & 70.0 & 69.7 & 41.1 & 65.6 & 50.3 & \textbf{69.0} & 59.1 ($-9.4$) \\
\bottomrule
\end{tabular}}%
\end{table}

\subsection{Scaling Experiments}
\label{app:scaling}

\emph{The complete scaling controls reproduce the main trend.}
Across both model-scale sweeps, SST preserves substantially more prior capability than unconstrained tuning while remaining competitive on the target task, and the retention advantage strengthens with model size. The matched-budget data comparison shows the same separation under both repeated and novel exposure (Figure~\ref{fig:scaling-data} in the main text): SFT drifts steadily, whereas SST remains stable and continues learning. Tables~\ref{tab:scale-e5} and~\ref{tab:scale} report the exact values.


\begin{table}[htbp]
\centering
\caption{\textbf{Five-epoch model-scale results underlying Figure~\ref{fig:scaling-size}.}
Science fine-tuning with the same configuration at each size. LM avg is the five-benchmark
average and $\Delta$ is relative to each model's own base. Bold entries mark column maxima among non-base recipes within each model size.}
\label{tab:scale-e5}
\small
\setlength{\tabcolsep}{3pt}
\resizebox{\textwidth}{!}{%
\begin{tabular}{llcccccccc}
\toprule
Model & Method & ID & GPQA & MMLU-Pro & GSM8K & HumanEval & IFEval & LM avg & $\Delta$ \\
\midrule
\multirow{4}{*}{Qwen3-14B}
 & Base & 69.4 & 62.1 & 75.5 & 95.6 & 93.9 & 85.8 & 82.6 & $+0.0$ \\
 & SFT & 78.0 & 39.9 & 48.9 & 72.4 & 36.6 & 26.8 & 44.9 & $-37.7$ \\
 & SFT+SST & 86.5 & \textbf{63.6} & 75.4 & 94.5 & \textbf{95.1} & 79.7 & \textbf{81.7} & $\mathbf{-0.9}$ \\
 & SFT+SST+R10 & \textbf{88.1} & 62.6 & \textbf{76.2} & \textbf{94.7} & 89.6 & \textbf{82.1} & 81.0 & $-1.6$ \\
\midrule
\multirow{4}{*}{Qwen3-8B}
 & Base & 60.2 & 60.1 & 73.0 & 94.8 & 91.5 & 85.6 & 81.0 & $+0.0$ \\
 & SFT & 76.1 & 35.4 & 48.7 & 81.0 & 36.0 & 34.2 & 47.1 & $-33.9$ \\
 & SFT+SST & \textbf{85.4} & \textbf{61.1} & 72.3 & 94.4 & 91.5 & \textbf{81.1} & \textbf{80.1} & $\mathbf{-0.9}$ \\
 & SFT+SST+R10 & 85.3 & 58.6 & \textbf{74.1} & \textbf{94.5} & \textbf{92.7} & 80.2 & 80.0 & $-1.0$ \\
\midrule
\multirow{4}{*}{Qwen3-4B}
 & Base & 67.9 & 55.1 & 70.0 & 94.0 & 92.1 & 82.8 & 78.8 & $+0.0$ \\
 & SFT & 73.4 & 32.3 & 44.2 & 78.3 & 29.9 & 27.0 & 42.3 & $-36.5$ \\
 & SFT+SST & 80.4 & 51.5 & 67.7 & \textbf{94.1} & \textbf{90.2} & 72.5 & 75.2 & $-3.6$ \\
 & SFT+SST+R10 & \textbf{84.3} & \textbf{53.0} & \textbf{68.7} & 93.9 & 86.0 & \textbf{75.6} & \textbf{75.4} & $\mathbf{-3.4}$ \\
\midrule
\multirow{4}{*}{Qwen3-1.7B}
 & Base & 54.1 & 38.4 & 58.6 & 89.3 & 83.5 & 72.3 & 68.4 & $+0.0$ \\
 & SFT & 69.5 & 27.3 & 38.6 & 72.4 & 23.8 & 27.9 & 38.0 & $-30.4$ \\
 & SFT+SST & 67.1 & 32.3 & 54.3 & \textbf{87.1} & 67.7 & \textbf{60.4} & 60.4 & $-8.0$ \\
 & SFT+SST+R10 & \textbf{70.4} & \textbf{34.8} & \textbf{56.0} & 86.6 & \textbf{71.3} & 58.6 & \textbf{61.5} & $\mathbf{-6.9}$ \\
\bottomrule
\end{tabular}}
\end{table}

\begin{table}[htbp]
\centering
\caption{\textbf{Two-epoch model-scale sweep on Science.} $\Delta$ is relative to each model's own baseline. Cells marked --- were not produced; bold entries mark column maxima among non-base recipes within each model size.}
\label{tab:scale}
\small
\setlength{\tabcolsep}{3pt}
\resizebox{\textwidth}{!}{%
\begin{tabular}{llcccccccc}
\toprule
Model & Method & ID & GPQA & MMLU-Pro & GSM8K & HumanEval & IFEval & LM avg & $\Delta$ \\
\midrule\midrule
\multirow{5}{*}{Qwen3-32B}
  & Baseline & 68.2 & 64.6 & 73.3 & 95.2 & 94.5 & 85.2 & 82.6 & $+0.0$ \\
\cmidrule{2-10}
  & SFT & 83.6 & 50.5 & 62.4 & 93.6 & 80.5 & 48.2 & 67.0 & $-15.6$ \\
  & SFT+SST & 84.1 & 62.6 & 73.5 & 94.8 & \textbf{97.0} & 79.9 & 81.6 & $-1.0$ \\
  & SFT+SST+R100 & 85.6 & 58.6 & 73.8 & \textbf{95.2} & 95.1 & \textbf{81.0} & 80.7 & $-1.9$ \\
  & SFT+SST+R10 & \textbf{87.3} & \textbf{65.7} & \textbf{74.3} & 95.1 & 94.5 & 79.1 & \textbf{81.7} & $\mathbf{-0.9}$ \\
\midrule\midrule
\multirow{9}{*}{Qwen3-14B}
  & Baseline & 69.4 & 62.1 & 75.5 & 95.6 & 93.9 & 85.8 & 82.6 & $+0.0$ \\
\cmidrule{2-10}
  & SFT & 82.3 & 50.0 & 64.7 & 93.9 & 64.0 & 54.5 & 65.4 & $-17.2$ \\
  & SFT+SST & 84.0 & \textbf{66.2} & 75.5 & \textbf{94.8} & 93.9 & 82.6 & \textbf{82.6} & $\mathbf{+0.0}$ \\
  & SFT+SST+R100 & 85.0 & 59.6 & \textbf{75.9} & 94.5 & 94.5 & \textbf{84.8} & 81.9 & $-0.7$ \\
  & SFT+SST+R10 & 86.3 & 60.6 & 75.6 & \textbf{94.8} & 92.1 & 81.0 & 80.8 & $-1.8$ \\
\cmidrule{2-10}
  & SDFT & 85.2 & 38.9 & 60.2 & 91.0 & 77.4 & 50.6 & 63.6 & $-19.0$ \\
  & SDFT+SST & 83.7 & 53.0 & 71.1 & 94.4 & \textbf{95.7} & 83.7 & 79.6 & $-3.0$ \\
  & SDFT+SST+R100 & 87.9 & 53.5 & 73.5 & 94.7 & 94.5 & 84.1 & 80.1 & $-2.5$ \\
  & SDFT+SST+R10 & \textbf{88.1} & 55.1 & 73.9 & 94.4 & 94.5 & 82.8 & 80.1 & $-2.5$ \\
\midrule\midrule
\multirow{9}{*}{Qwen3-8B}
  & Baseline & 60.2 & 60.1 & 73.0 & 94.8 & 91.5 & 85.6 & 81.0 & $+0.0$ \\
\cmidrule{2-10}
  & SFT & 77.6 & 38.4 & 62.3 & 93.9 & 36.6 & 46.2 & 55.5 & $-25.5$ \\
  & SFT+SST & 80.1 & 55.1 & 72.0 & 94.2 & 92.7 & 80.4 & 78.9 & $-2.1$ \\
  & SFT+SST+R100 & 81.7 & 58.1 & 71.9 & 94.8 & 91.5 & 82.1 & 79.7 & $-1.3$ \\
  & SFT+SST+R10 & 84.0 & \textbf{58.6} & \textbf{74.4} & 94.9 & 92.7 & 80.4 & 80.2 & $-0.8$ \\
\cmidrule{2-10}
  & SDFT & \textbf{88.1} & 42.4 & 62.5 & 92.9 & 61.6 & 25.3 & 56.9 & $-24.1$ \\
  & SDFT+SST & 82.9 & 53.0 & 72.5 & 94.8 & 92.1 & \textbf{85.0} & 79.5 & $-1.5$ \\
  & SDFT+SST+R100 & 83.4 & 57.1 & 73.1 & \textbf{95.1} & 93.3 & 84.7 & \textbf{80.7} & $\mathbf{-0.3}$ \\
  & SDFT+SST+R10 & 85.8 & 56.6 & 70.2 & 93.6 & \textbf{95.7} & 84.1 & 80.0 & $-1.0$ \\
\midrule\midrule
\multirow{9}{*}{Qwen3-4B}
 & Baseline & 67.9 & 55.1 & 70.0 & 94.0 & 92.1 & 82.8 & 78.8 & $+0.0$ \\
\cmidrule{2-10}
 & SFT & 76.5 & 38.9 & 61.1 & 92.1 & 24.4 & 42.1 & 51.7 & $-27.1$ \\
 & SFT+SST & 78.1 & 53.0 & 68.2 & \textbf{93.8} & \textbf{93.9} & 79.9 & \textbf{77.8} & $\mathbf{-1.0}$ \\
 & SFT+SST+R100 & 78.7 & \textbf{55.1} & 67.8 & 93.6 & 92.7 & 79.5 & 77.7 & $-1.1$ \\
 & SFT+SST+R10 & 80.6 & 53.5 & \textbf{68.6} & 93.5 & 88.4 & 79.5 & 76.7 & $-2.1$ \\
\cmidrule{2-10}
 & SDFT & 81.9 & 34.8 & 51.5 & 84.1 & 58.5 & 42.3 & 54.2 & $-24.6$ \\
 & SDFT+SST & 81.1 & 42.4 & 61.7 & 92.4 & 89.6 & 73.8 & 72.0 & $-6.8$ \\
 & SDFT+SST+R100 & 83.9 & 44.9 & 65.0 & 92.6 & 92.7 & 77.8 & 74.6 & $-4.2$ \\
 & SDFT+SST+R10 & \textbf{86.1} & 48.5 & 67.9 & 92.7 & 90.9 & \textbf{80.2} & 76.0 & $-2.8$ \\
\midrule\midrule
\multirow{9}{*}{Qwen3-1.7B}
 & Baseline & 54.1 & 38.4 & 58.6 & 89.3 & 83.5 & 72.3 & 68.4 & $+0.0$ \\
\cmidrule{2-10}
 & SFT & 66.2 & 29.3 & 51.5 & 85.2 & 48.2 & 34.8 & 49.8 & $-18.6$ \\
 & SFT+SST & 65.3 & 31.8 & \textbf{55.8} & 87.6 & \textbf{75.0} & 63.6 & 62.8 & $-5.6$ \\
 & SFT+SST+R100 & 65.1 & 33.3 & 55.1 & \textbf{88.2} & 72.6 & \textbf{65.4} & \textbf{62.9} & $\mathbf{-5.5}$ \\
 & SFT+SST+R10 & 63.8 & \textbf{37.4} & 54.3 & 86.8 & 69.5 & 61.0 & 61.8 & $-6.6$ \\
\cmidrule{2-10}
 & SDFT & 75.4 & 27.3 & 36.0 & 57.2 & 41.5 & 25.7 & 37.5 & $-30.9$ \\
 & SDFT+SST & 65.7 & 23.2 & 40.4 & 80.3 & 69.5 & 61.7 & 55.0 & $-13.4$ \\
 & SDFT+SST+R100 & 72.7 & 27.3 & 45.5 & 82.6 & 72.6 & 64.1 & 58.4 & $-10.0$ \\
 & SDFT+SST+R10 & \textbf{76.1} & 30.3 & 48.6 & 84.2 & 74.4 & 61.7 & 59.8 & $-8.6$ \\
\bottomrule
\end{tabular}}\end{table}
\subsection{Qwen3-8B thinking: full result tables}
\label{app:qwen3-full}

\emph{The full per-domain sweep.} Across Science, Finance, Pathology, and Medicine, unconstrained recipes learn the target task but broadly degrade prior capability, while adding SST keeps the model much closer to its baseline while retaining strong target-task performance. Table~\ref{tab:qwen3-main} reports the exact values for every recipe and benchmark.


\begin{table}[htbp]
\centering
\caption{Qwen3-8B (thinking mode). Methods are grouped by recipe. Bold entries mark column maxima among non-base recipes within each domain.
$\Delta$ = LM avg $-$ baseline ($81.0$).}
\label{tab:qwen3-main}
\small
\setlength{\tabcolsep}{2.6pt}
\resizebox{0.75\textwidth}{!}{%
\begin{tabular}{llcccccccc}
\toprule
Domain & Method & ID & GPQA & MMLU-Pro & GSM8K & HumanEval & IFEval & LM avg & $\Delta$ \\
\midrule\midrule
\multirow{18}{*}{Science}
 & Baseline             & 60.2 & 60.1 & 73.0 & 94.8 & 91.5 & 85.6 & 81.0 & $+0.0$ \\
\cmidrule{2-10}
 & SFT                  & 77.6 & 38.4 & 62.3 & 93.9 & 36.6 & 46.2 & 55.5 & $-25.5$ \\
 & SFT+SST & 80.1 & 57.6 & 73.1 & \textbf{95.1} & 92.1 & 82.1 & 80.0 & $-1.0$ \\
 & SFT+SST+R100 & 81.7 & 57.6 & 73.1 & \textbf{95.1} & 92.1 & 82.1 & 80.0 & $-1.0$ \\
 & SFT+SST+R10  & 84.0 & 58.6 & \textbf{74.4} & 94.9 & 92.7 & 80.4 & 80.2 & $-0.8$ \\
\cmidrule{2-10}
 & SDFT                 & \textbf{88.1} & 42.4 & 62.5 & 92.9 & 61.6 & 25.3 & 57.0 & $-24.0$ \\
 & SDFT+SST& 82.9 & 53.0 & 72.5 & 94.8 & 92.1 & \textbf{85.0} & 79.5 & $-1.5$ \\
 & SDFT+SST+R100& 83.4 & 57.1 & 73.1 & \textbf{95.1} & 93.3 & 84.7 & \textbf{80.7} & $\mathbf{-0.3}$ \\
 & SDFT+SST+R10 & 85.8 & 56.6 & 70.2 & 93.6 & \textbf{95.7} & 84.1 & 80.0 & $-1.0$ \\
\cmidrule{2-10}
 & SFT+Muon             & 80.2 & 51.0 & 64.4 & 93.8 & 42.1 & 46.3 & 59.5 & $-21.5$ \\
 & SFT+Muon+SST & 80.9 & \textbf{59.1} & 72.0 & 94.5 & 91.5 & 81.3 & 79.7 & $-1.3$ \\
 & SFT+Muon+SST+R10 & 82.1 & 55.6 & 73.3 & 95.0 & 90.2 & 82.8 & 79.4 & $-1.6$ \\
\cmidrule{2-10}
 & SFT+ReInvoke         & 68.4 & 52.5 & 64.6 & 91.1 & 76.8 & 66.4 & 70.3 & $-10.7$ \\
 & OPLoRA ($k{=}0.99d$)  & 80.8 & 53.5 & 72.2 & 94.9 & 90.9 & 84.7 & 79.2 & $-1.8$ \\
\midrule\midrule
\multirow{14}{*}{FinQA}
 & Baseline & 70.5 & 60.1 & 73.0 & 94.8 & 91.5 & 85.6 & 81.0 & $+0.0$ \\
\cmidrule{2-10}
 & SFT & 72.3 & 10.1 & 15.5 & 43.9 & 15.9 & 33.1 & 23.7 & $-57.3$ \\
 & SFT+SST & 73.0 & \textbf{62.6} & \textbf{74.1} & 94.8 & \textbf{93.9} & 86.0 & \textbf{82.3} & $\mathbf{+1.3}$ \\
 & SFT+SST+R100 & 73.1 & 53.5 & 73.9 & 94.7 & 93.3 & 84.8 & 80.1 & $-0.9$ \\
 & SFT+SST+R10 & \textbf{75.3} & 40.4 & 68.3 & 94.6 & 89.0 & \textbf{86.3} & 75.7 & $-5.3$ \\
\cmidrule{2-10}
 & SDFT & 67.9 & 27.3 & 34.2 & 88.1 & 89.0 & 41.4 & 56.0 & $-25.0$ \\
 & SDFT+SST & 64.1 & 57.6 & 73.9 & 95.1 & 93.3 & 86.1 & 81.2 & $+0.2$ \\
 & SDFT+SST+R100 & 67.6 & 55.1 & 72.3 & 94.5 & \textbf{93.9} & 84.3 & 80.0 & $-1.0$ \\
 & SDFT+SST+R10 & 74.1 & 51.0 & 65.5 & 95.2 & \textbf{93.9} & 80.8 & 77.3 & $-3.7$ \\
\cmidrule{2-10}
 & SFT+Muon & 74.9 & 16.2 & 16.9 & 63.7 & 42.1 & 45.8 & 36.9 & $-44.1$ \\
 & SFT+Muon+SST & 71.4 & 60.1 & 73.4 & \textbf{95.3} & 91.5 & 86.0 & 81.2 & $+0.2$ \\
 & SFT+Muon+SST+R10 & 74.3 & 36.9 & 67.9 & 94.1 & 92.1 & 85.0 & 75.2 & $-5.8$ \\
\cmidrule{2-10}
 & SFT+ReInvoke & 72.2 & 9.6 & 15.3 & 43.7 & 16.5 & 27.4 & 22.5 & $-58.5$ \\
 & OPLoRA ($k{=}0.99d$) & 71.5 & 32.8 & 63.4 & 93.6 & 89.0 & 82.8 & 72.3 & $-8.7$ \\
\midrule\midrule
\multirow{18}{*}{Pathology}
 & Baseline             & 78.3 & 60.1 & 73.0 & 94.8 & 91.5 & 85.6 & 81.0 & $+0.0$ \\
\cmidrule{2-10}
 & SFT                  & 80.4 & 28.3 & 48.6 & 84.5 &  0.6 & 11.6 & 34.7 & $-46.3$ \\
 & SFT+SST & 77.3 & 53.0 & 67.0 & 94.4 & 90.9 & 81.0 & 77.2 & $-3.8$ \\
 & SFT+SST+R100 & 77.6 & 55.1 & 68.6 & 94.8 & 93.3 & 81.0 & 78.5 & $-2.5$ \\
 & SFT+SST+R10  & 82.1 & 39.9 & 65.6 & 91.4 & 92.7 & 72.1 & 72.3 & $-8.7$ \\
\cmidrule{2-10}
 & SDFT                 & \textbf{84.9} & 49.0 & 67.5 & 94.2 & 76.2 & 22.9 & 62.0 & $-19.0$ \\
 & SDFT+SST& 82.0 & 56.1 & 72.5 & 94.6 & \textbf{95.1} & 82.6 & 80.2 & $-0.8$ \\
 & SDFT+SST+R100& 80.3 & \textbf{59.1} & \textbf{72.8} & 95.1 & 92.1 & \textbf{83.4} & \textbf{80.5} & $\mathbf{-0.5}$ \\
 & SDFT+SST+R10 & 83.2 & 53.0 & 72.0 & \textbf{95.6} & 93.9 & 82.3 & 79.4 & $-1.6$ \\
\cmidrule{2-10}
 & SFT+Muon             & 79.3 & 36.9 & 52.2 & 89.1 &  1.8 & 14.2 & 38.8 & $-42.2$ \\
 & SFT+Muon+SST & 77.0 & \textbf{59.1} & 70.0 & 95.1 & 94.5 & 81.5 & 80.0 & $-1.0$ \\
 & SFT+Muon+SST+R10 & 78.4 & 51.0 & 64.0 & 95.4 & 93.3 & 78.7 & 76.5 & $-4.5$ \\
\cmidrule{2-10}
 & SFT+ReInvoke         & 77.6 & 49.5 & 69.5 & 93.9 & 85.4 & 70.2 & 73.7 & $-7.3$ \\
 & OPLoRA ($k{=}0.99d$)  & 76.2 & 38.9 & 58.4 & 93.6 & 91.5 & 71.3 & 70.7 & $-10.3$ \\
\midrule\midrule
\multirow{18}{*}{Medicine}
 & Baseline             & 74.8 & 60.1 & 73.0 & 94.8 & 91.5 & 85.6 & 81.0 & $+0.0$ \\
\cmidrule{2-10}
 & SFT                  & 72.9 & 33.3 & 49.0 & 87.6 &  0.0 & 11.6 & 36.3 & $-44.7$ \\
 & SFT+SST & 72.4 & 56.6 & 68.8 & 94.8 & 93.3 & 81.1 & 78.9 & $-2.1$ \\
 & SFT+SST+R100 & 74.3 & 55.1 & 68.1 & 94.7 & 93.3 & 80.2 & 78.3 & $-2.7$ \\
 & SFT+SST+R10  & 76.6 & 46.0 & 65.9 & 86.8 & 87.2 & 70.2 & 71.2 & $-9.8$ \\
\cmidrule{2-10}
 & SDFT                 & \textbf{79.6} & 59.6 & 70.1 & 93.9 & 82.9 & 25.1 & 66.3 & $-14.7$ \\
 & SDFT+SST& 74.8 & 58.1 & \textbf{72.5} & 94.8 & 92.7 & 85.2 & 80.7 & $-0.3$ \\
 & SDFT+SST+R100& 76.3 & \textbf{60.6} & \textbf{72.5} & \textbf{95.1} & \textbf{94.5} & \textbf{85.4} & \textbf{81.6} & $\mathbf{+0.6}$ \\
 & SDFT+SST+R10 & 77.5 & 52.5 & 72.1 & 94.1 & 93.9 & 82.3 & 79.0 & $-2.0$ \\
\cmidrule{2-10}
 & SFT+Muon             & 71.6 & 29.8 & 49.1 & 86.1 &  0.6 & 21.7 & 37.5 & $-43.5$ \\
 & SFT+Muon+SST & 72.8 & 54.5 & 70.0 & 94.9 & 90.9 & 80.8 & 78.2 & $-2.8$ \\
 & SFT+Muon+SST+R10 & 75.0 & 52.5 & 63.8 & 94.8 & 92.1 & 78.9 & 76.4 & $-4.6$ \\
\cmidrule{2-10}
 & SFT+ReInvoke         & 71.6 & 55.6 & 69.4 & 94.4 & 86.6 & 70.6 & 75.3 & $-5.7$ \\
 & OPLoRA ($k{=}0.99d$)  & 70.8 & 34.3 & 59.5 & 93.0 & 92.1 & 71.0 & 70.0 & $-11.0$ \\
\bottomrule
\end{tabular}}
\end{table}

\subsection{Policy drift across model sizes}
\label{app:kl-scale}

\begin{figure}[htbp]
\centering
\includegraphics[width=0.72\textwidth]{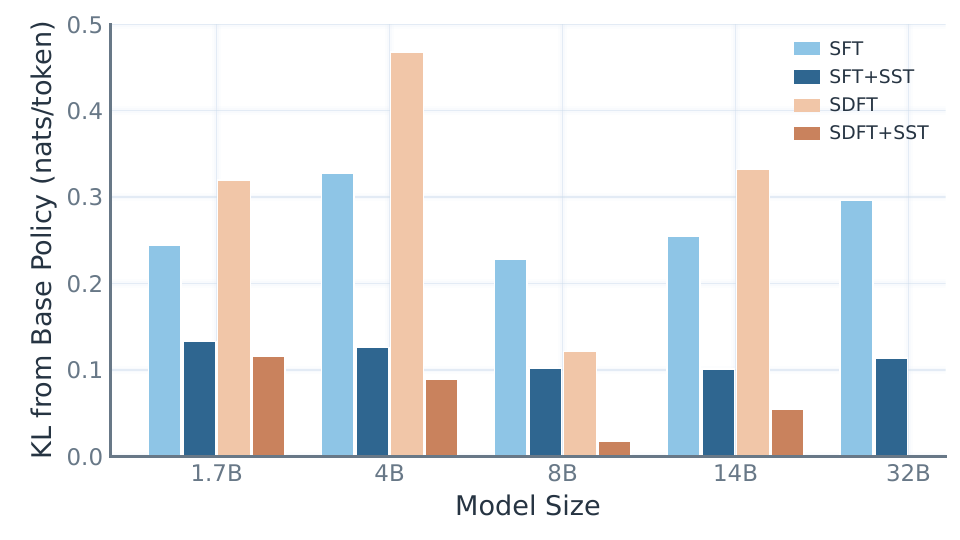}
\caption{Policy drift at the final fine-tuned checkpoint across Qwen3 sizes. Light bars are the recipe alone,
full-color bars their $+$ SST version (no-refresh).}
\label{fig:kl-scale-bars}
\end{figure}

\emph{SST reduces policy drift at every model scale.} Figure~\ref{fig:kl-scale-bars} reports WikiText-2 KL divergence at the final checkpoint, with each model scored against its own pretrained base. Across sizes, SST remains substantially closer to the base distribution than SFT, and the separation persists at the largest scale. The figure gives the exact drift values for each model.

\subsection{Toy study: full specification}
\label{app:toy}

\emph{Applied-update projection is the only constraint that is both exact and stable.} The controlled setup below isolates where the projection is applied. All schemes learn the toy task, but only projecting the optimizer's applied update eliminates protected-subspace leakage without substantially changing the held-out capability.

\paragraph{Model, optimizer, targets.} Qwen2.5-0.5B, fine-tuned with plain AdamW at learning
rate $1{\times}10^{-4}$ and weight decay $0$ (unless stated). Each scheme's projection is applied to
the standard target linears---attention $q,k,v,o$ and MLP $gate,up,down$.

\paragraph{Task ($8$ fixed examples).} Next-token cross-entropy on the completion tokens only (prompt
tokens masked), on eight short prompt$\to$completion pairs: \texttt{("The capital of France is",
" Paris.")}, \texttt{("2 plus 3 equals", " 5.")}, \texttt{("The opposite of hot is", " cold.")},
\texttt{("Water is made of hydrogen and", " oxygen.")}, \texttt{("The first month of the year is",
" January.")}, \texttt{("A group of wolves is called a", " pack.")}, \texttt{("The chemical symbol for
gold is", " Au.")}, \texttt{("Seven times eight is", " 56.")}. The task is trivially learnable, so
every scheme drives task loss to ${\approx}0$; the discriminator is the forgetting on out of domain tasks.

\paragraph{Capability probe (held-out).} Mean per-token NLL on four generic sentences the model never
trains on; \textbf{Probe}$\Delta = \mathrm{NLL}_\text{after}-\mathrm{NLL}_\text{before}$ (base-model
probe NLL ${\approx}2.81$ nats/token). The probe sentences: \emph{``The history of the Roman Empire
spans over a thousand years of political change.''}; \emph{``Photosynthesis converts carbon dioxide
and water into glucose using sunlight.''}; \emph{``In distributed systems, consensus algorithms ensure
agreement despite failures.''}; \emph{``The novel explores themes of memory, loss, and the passage of
time.''}

\paragraph{The three enforcement schemes.} With projector $\Pi$ (Eq.~\ref{eq:proj}), Adam moments
$\hat m,\hat v$, and preconditioned raw step $\hat m/(\sqrt{\hat v}+\epsilon)$:
\begin{itemize}
\item \textbf{Project gradient}: $g\leftarrow\Pi(g)$; then $\hat m,\hat v \leftarrow \text{Adam}(g)$;
$W\leftarrow W-\text{lr}\cdot\hat m/(\sqrt{\hat v}+\epsilon)$. Here $\hat v$ sees only free-subspace
energy, and the applied step $D\,\Pi(\hat m)$ with $D=\mathrm{diag}(1/(\sqrt{\hat v}+\epsilon))$
re-enters the protected space because $D$ and $\Pi$ do not commute.
\item \textbf{Project the applied update}: $\hat m,\hat v\leftarrow\text{Adam}(g)$ on the \emph{full}
gradient; $W_\text{pre}\leftarrow W$; $W\leftarrow W-\text{lr}\cdot\hat m/(\sqrt{\hat v}+\epsilon)$;
$W\leftarrow W_\text{pre}+\Pi(W-W_\text{pre})$. The update is confined to the low singular value subspace because projection is last, and
the free step keeps the scale induced by the full gradient.
\item \textbf{Re-project cumulative change}: $g\leftarrow\Pi(g)$; take the Adam step; then set
$W\leftarrow W_\text{ref}+\Pi(W-W_\text{ref})$. Note that $W\leftarrow W_\text{ref}+\Pi(W-W_\text{ref})$
is equivalent to $W_t+\Pi(\delta_t)$, with $\delta_t$ the step just applied, since every previously
applied change already lies in the free subspace. The update lies in the correct subspace, but the step is based on
projected-gradient moments and is therefore mis-scaled.
\end{itemize}

\paragraph{The full procedure.} Algorithm~\ref{alg:sst} details SST's implementation, using the second
of these three schemes. A superscript $\ell$ indexes one constrained matrix, so $W_t^\ell$ is that
matrix at step $t$ and plays the role of $W_\text{pre}$ above. The optimizer itself is never
modified. It receives the full gradient, updates its moments, and proposes an ordinary step; SST
then replaces the part of that step which lands on a constrained matrix with its free-to-free block.
Parameters outside $\mathcal I$, such as embeddings and layer norms, are left to the optimizer.

\begin{algorithm}[htbp]
\caption{Subspace Tuning (SST). The optimizer runs untouched; SST edits only the update it
produces, keeping the free-to-free block of Eq.~\ref{eq:proj} and discarding the rest.}
\label{alg:sst}
\begin{algorithmic}[1]
\Require Constrained matrices $\mathcal I$ (attention $q,k,v,o$ and MLP gate, up, down) with
         pretrained weights $W^\ell$ of shape $d_\text{out}^\ell\times d_\text{in}^\ell$; rank
         fraction $r$, the share of singular directions held fixed; refresh interval $K$ in
         optimizer steps; optimizer $\mathcal O$ with state $s$; loss $\mathcal L$ with minibatches
         $\mathcal B_t$; step budget $T$.
\Statex
\Statex \textit{Setup: protect the leading singular subspace of all matrices in $\mathcal I$.}
\ForAll{$\ell\in\mathcal I$}
    \State $W_\text{ref}^\ell\gets W^\ell$
    \State $k_\ell\gets\left\lfloor r\min(d_\text{out}^\ell,d_\text{in}^\ell)\right\rfloor$
           \Comment{number of directions to protect}
    \State $U_k^\ell,V_k^\ell\gets$ top-$k_\ell$ left and right singular vectors of $W_\text{ref}^\ell$
    \State $Q_U^\ell\gets I-U_k^\ell(U_k^\ell)^\top,\quad
           Q_V^\ell\gets I-V_k^\ell(V_k^\ell)^\top$
           \Comment{the free complements}
\EndFor
\Statex
\Statex \textit{Training: take the ordinary step, then project the part that reaches the weights.}
\For{$t=0,\ldots,T-1$}
    \State $g_t\gets\nabla_W\mathcal L(W_t;\mathcal B_t)$
           \Comment{full gradient}
    \State $(\widetilde W_{t+1},s_{t+1})\gets\mathcal O(W_t,g_t,s_t)$
           \Comment{ordinary step}
    \State $W_{t+1}\gets\widetilde W_{t+1}$
    \ForAll{$\ell\in\mathcal I$}
        \State $\Delta W^\ell\gets\widetilde W_{t+1}^\ell-W_t^\ell$
               \Comment{the optimizer update}
        \State $W_{t+1}^\ell\gets W_t^\ell+
               Q_U^\ell\,\Delta W^\ell\,Q_V^\ell$
               \Comment{apply $\Pi(\Delta W^\ell)$ instead, Eq.~\ref{eq:proj}}
    \EndFor
    \If{$K<\infty$ and $(t+1)\bmod K=0$}
        \ForAll{$\ell\in\mathcal I$}
            \State $W_\text{ref}^\ell\gets W_{t+1}^\ell$
                   \Comment{re-compute projection with current weights}
            \State $U_k^\ell,V_k^\ell\gets$ top-$k_\ell$ singular vectors of $W_\text{ref}^\ell$
            \State $Q_U^\ell\gets I-U_k^\ell(U_k^\ell)^\top,\quad
                   Q_V^\ell\gets I-V_k^\ell(V_k^\ell)^\top$
        \EndFor
    \EndIf
\EndFor
\State \Return $W_T$
\end{algorithmic}
\end{algorithm}


\subsection{Comparison with subspace-constrained baselines}
\label{app:baselines}

The closest baselines are Sculpting \citep{nayak2025sculpting} and OPLoRA
\citep{xiong2025oplora}. Both restrict updates to low-energy directions through a trainable
factorisation; SST instead projects the applied full-parameter update. We use the official
hyperparameters reported in each baseline's paper and otherwise keep the data, training budget,
and evaluation harness fixed.

\emph{SST preserves more capability than factorised subspace baselines.} Table~\ref{tab:baselines-4dom} shows that Sculpting and OPLoRA reduce forgetting relative to SFT, but remain below no-refresh SST on prior capability across the four-domain comparison. The gap is modest on Science and becomes clearer on Finance and the medical domains. Refresh shifts SST toward stronger task learning at some cost to retention. The table reports the exact domain-level endpoints.

\begin{figure}[htb]
\centering
\includegraphics[width=\textwidth]{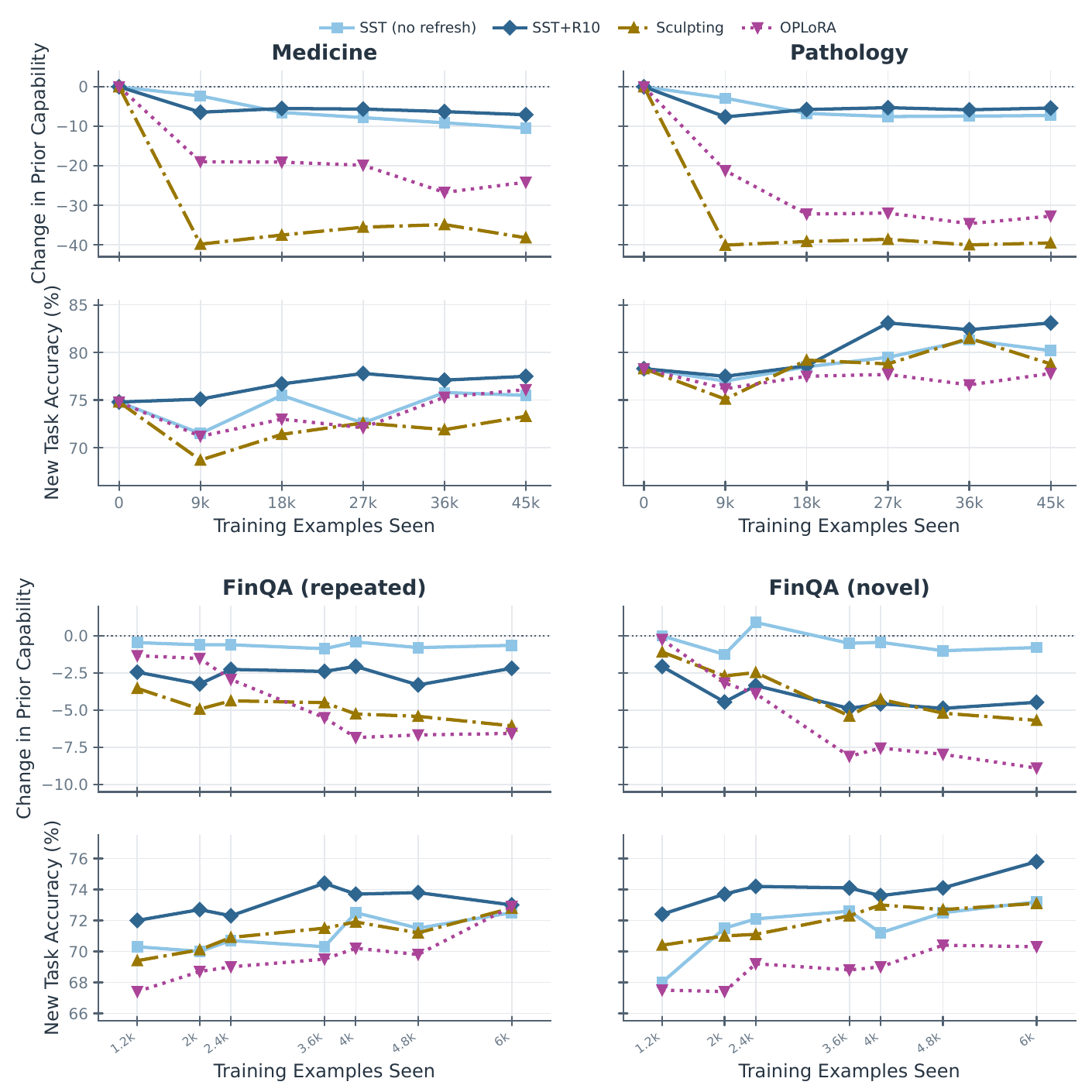}
\caption{\textbf{Data scaling against subspace-constrained baselines on Qwen3-8B.}
The top two rows show Medicine (left) and Pathology (right), each using $9{,}000$ unique examples
repeated for five epochs. The bottom two rows show FinQA with matched $6{,}000$-example exposure:
$2$k unique traces repeated for three epochs (left) or $6$k unique traces seen once (right).
Within each pair, the upper panels show change in the five-benchmark prior-capability average from
the base value of $81.0$, and the lower panels show target accuracy. Methods share the data and
evaluation schedule within each setting. Curves compare no-refresh SST, SST+R10, Sculpting, and
OPLoRA; SFT is omitted to keep the constrained trajectories legible.}
\label{fig:finqa-datascale}
\end{figure}

\paragraph{Data scaling across Medicine, Pathology, and FinQA.}
Figure~\ref{fig:finqa-datascale} combines the repeated-data Medicine and Pathology sweeps with the
matched-exposure FinQA comparison. Medicine and Pathology each repeat $9{,}000$ examples for five
epochs ($45{,}000$ examples total), with evaluation every $9{,}000$ examples. FinQA compares
$2{,}000$ unique examples repeated for three epochs against $6{,}000$ unique examples seen once;
both arms therefore contain $6{,}000$ examples of exposure and use the same held-out $1{,}000$
examples.

\emph{SST retains capability while continuing to learn.} Both SST variants preserve substantially
more prior capability than Sculpting and OPLoRA on Medicine and Pathology. On FinQA, no-refresh
SST remains closest to the base model, while R10 reaches higher target accuracy in both exposure
regimes. SFT is omitted so the constrained trajectories remain legible.

\clearpage

\begin{table}[htb]
\centering
\caption{Comparison with subspace-constrained baselines using the official method-specific settings
from their papers. All methods use the same datasets and evaluation suite. ID is in-domain accuracy and
LM is the five-benchmark capability average (base $81.0$). Bold entries mark column maxima among non-base recipes.}
\label{tab:baselines-4dom}
\small
\setlength{\tabcolsep}{4pt}
\begin{tabular}{lcccccccccc}
\toprule
& \multicolumn{2}{c}{Science} & \multicolumn{2}{c}{FinQA} & \multicolumn{2}{c}{Pathology}
& \multicolumn{2}{c}{Medicine} & \multicolumn{2}{c}{Mean} \\
\cmidrule(lr){2-3}\cmidrule(lr){4-5}\cmidrule(lr){6-7}\cmidrule(lr){8-9}\cmidrule(lr){10-11}
Method & ID & LM & ID & LM & ID & LM & ID & LM & ID & LM \\
\midrule
Baseline            & 60.2 & 81.0 & 70.5 & 81.0 & 78.3 & 81.0 & 74.8 & 81.0 & 71.0 & 81.0 \\
SFT                 & 77.6 & 55.5 & 72.3 & 23.7 & 80.4 & 34.7 & 72.9 & 36.3 & 75.8 & 37.6 \\
\midrule
Sculpting           & 83.2 & 77.2 & 74.5 & 57.7 & 79.4 & 43.5 & 74.6 & 41.3 & 77.9 & 54.9 \\
OPLoRA              & 78.5 & 79.1 & 72.8 & 61.7 & 77.4 & 51.6 & 70.0 & 60.0 & 74.7 & 63.1 \\
\midrule
SFT+SST      & 80.1 & 80.0 & 73.0 & \textbf{82.3} & 77.3 & 77.2 & 72.4 & \textbf{78.9} & 75.7 & \textbf{79.6} \\
SFT+SST+R100 & 81.7 & 80.0 & 73.1 & 80.1 & 77.6 & \textbf{78.5} & 74.3 & 78.3 & 76.7 & 79.2 \\
SFT+SST+R10  & \textbf{84.0} & \textbf{80.2} & \textbf{75.3} & 75.7 & \textbf{82.1} & 72.3 & \textbf{76.6} & 71.2 & \textbf{79.5} & 74.9 \\
\bottomrule
\end{tabular}
\end{table}

\subsection{KL regularisation does not prevent forgetting}
\label{app:kl-reg}

\begin{figure}[htb]
\centering
\includegraphics[width=\textwidth]{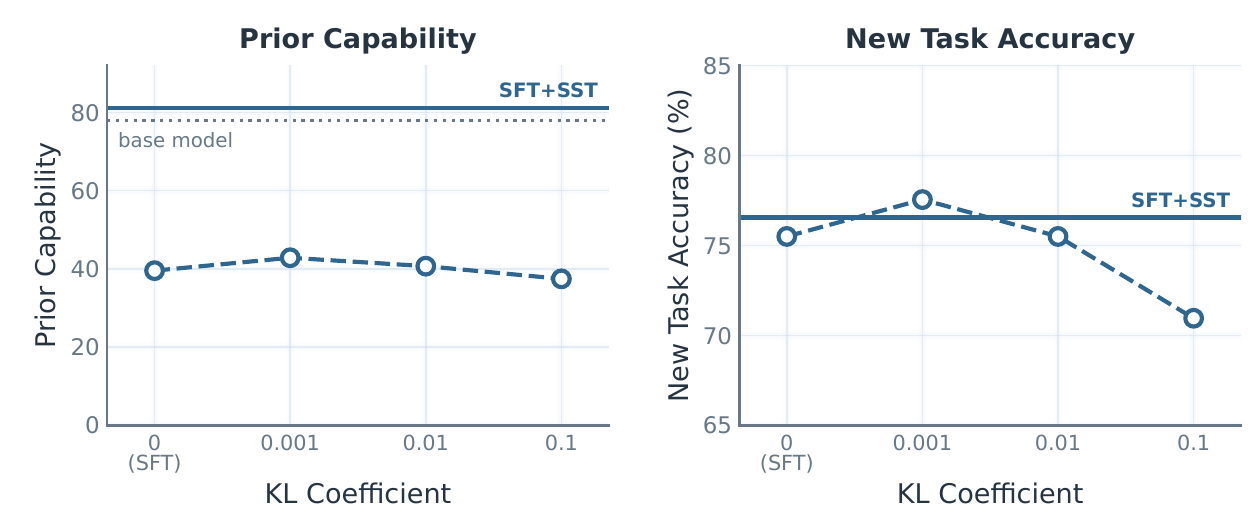}
\caption{\textbf{Reverse-KL regularisation across three coefficients.} Prior capability (left) and
in-domain accuracy (right) against SFT with KL regularization, averaged over FinQA and Science. The grey rule
is the un-finetuned base model and the solid blue rule is SST at $r{=}0.99$ on the same two tasks.}
\label{fig:kl-reg}
\end{figure}

\emph{Penalising the policy does not mitigate forgetting.} The natural alternative to constraining
\emph{where} an update may move is to leave the update unconstrained and instead penalise how far
the resulting policy drifts. We fine-tune Qwen3-8B on FinQA and on Science with
$\mathcal{L}=\mathrm{CE}+\lambda\,\mathrm{KL}(\pi_\theta\,\|\,\pi_\mathrm{base})$, a reverse KL over
the response tokens against a frozen copy of the base model, at
$\lambda\in\{0.001,0.01,0.1\}$, plus a $\lambda{=}0$ control on FinQA.
Table~\ref{tab:kl-reg} and Figure~\ref{fig:kl-reg} show that reverse-KL regularisation does not provide a reliable learning--retention trade-off. Weak regularisation leaves forgetting essentially unchanged, while stronger regularisation can reduce target-task learning without restoring prior capability. The effect also varies across tasks, unlike the consistent retention obtained by constraining the update subspace. The table reports the exact per-task and per-benchmark values.

\begin{table}[htb]
\centering
\caption{\textbf{SFT with reverse-KL regularisation, Qwen3-8B.} Prior capability is the mean of the
five held-out benchmarks shown. Bold entries mark column maxima among non-base recipes across the reported rows.}
\label{tab:kl-reg}
\small
\setlength{\tabcolsep}{4pt}
\resizebox{\textwidth}{!}{%
\begin{tabular}{llccccccc}
\toprule
Task & Method & ID & GPQA & MMLU-Pro & GSM8K & HumanEval & IFEval & Prior cap. \\
\midrule
--- & Base &  & 58.1 & 59.7 & 94.4 & 92.1 & 85.8 & 78.0 \\
\midrule\midrule
\multirow{4}{*}{FinQA}
 & $\lambda{=}0$ & 73.4 & 10.1 & 13.4 & 41.8 & 24.4 & 28.1 & 23.6 \\
 &  $\lambda{=}0.001$ & 73.4 & \phantom{0}7.6 & 14.2 & 43.5 & 24.4 & 26.4 & 23.2 \\
 &  $\lambda{=}0.01$  & 71.3 & \phantom{0}7.1 & 13.2 & 42.0 & 17.7 & 26.6 & 21.3 \\
 &  $\lambda{=}0.1$   & 66.0 & 13.6 & 15.8 & 14.2 & \phantom{0}0.0 & \phantom{0}9.4 & 10.6 \\
\midrule\midrule
\multirow{3}{*}{Science}
 & $\lambda{=}0.001$ & \textbf{81.7} & \textbf{49.5} & 54.4 & 93.3 & 60.4 & \textbf{55.1} & 62.5 \\
 & $\lambda{=}0.01$  & 79.7 & 46.5 & \textbf{55.0} & \textbf{93.5} & 50.6 & \textbf{55.1} & 60.1 \\
 &  $\lambda{=}0.1$   & 75.9 & 47.5 & 51.2 & 93.4 & \textbf{77.4} & 51.9 & \textbf{64.3} \\
\bottomrule
\end{tabular}}%
\end{table}

\subsection{Rank ablation on FinQA}
\label{app:rank-finqa}

\begin{figure}[htbp]
\centering
\includegraphics[width=\textwidth]{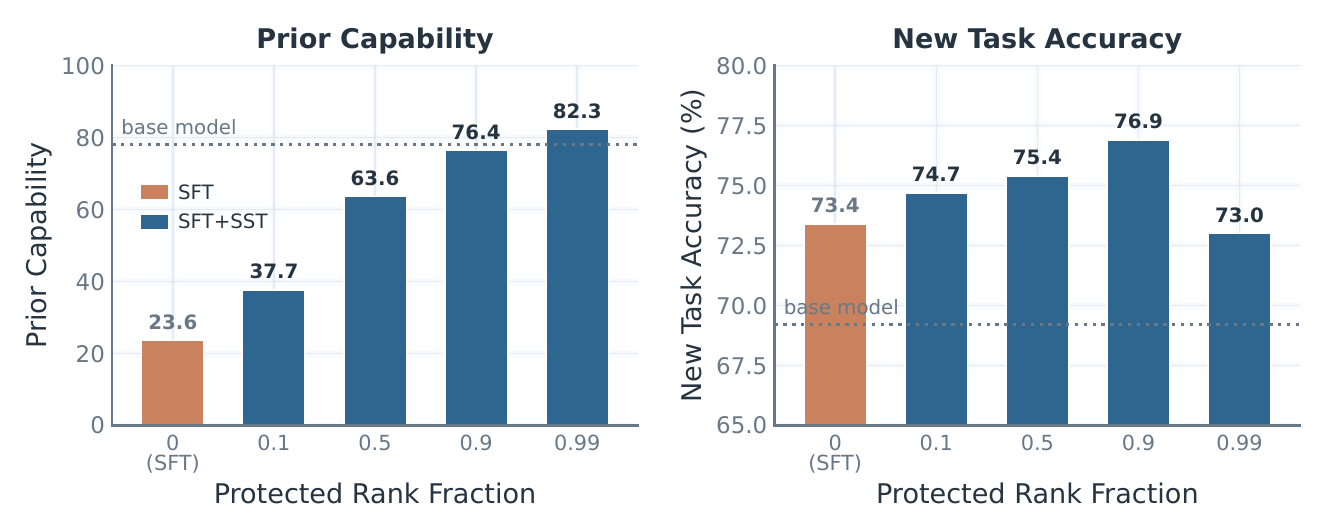}
\caption{\textbf{Protected rank fraction on FinQA.} Prior capability (left) and in-domain accuracy
(right) against the fraction of each weight's subspace held fixed, with plain SFT as the $r{=}0$
end of the same sweep. The grey rule is the un-finetuned base model. Where no KL coefficient recovers
prior capability (Figure~\ref{fig:kl-reg}), rank recovers it almost completely and in-domain accuracy
rises rather than falls.}
\label{fig:rank-finqa}
\end{figure}

\emph{The constraint maintains prior capability.} FinQA is the hardest case in
Section~\ref{app:kl-reg}: every KL coefficient leaves prior capability at or below plain SFT. We
repeat that setting with SST instead, sweeping the protected rank fraction
$r\in\{0.1,0.5,0.9\}$ with no refresh. Figure~\ref{fig:rank-finqa}
reports the sweep against the same base control and the same $\lambda{=}0$ SFT run.

\emph{Rank controls retention.} Figure~\ref{fig:rank-finqa} shows prior capability improving monotonically as a larger fraction of the weight subspace is protected. The held-out benchmarks move together, indicating a broad recovery rather than an average driven by one task.

\emph{Protecting more only impacts in-domain accuracy for high ranks.} In-domain accuracy also improves across the sweep until $r=0.99$, where the new task accuracy declines. Figure~\ref{fig:rank-finqa} shows the trajectory. Although a lower \(r\) permits adaptation along more directions, supervision with a dated projection matrix can disrupt general model behavior (e.g., collapse response length), ultimately hindering learning of the target distribution.


\subsection{Effect of Optimizer Hyperparameters on SST}
\label{app:finqa-optimizer-hparams}

\begin{figure}[H]
\centering
\includegraphics[width=\textwidth]{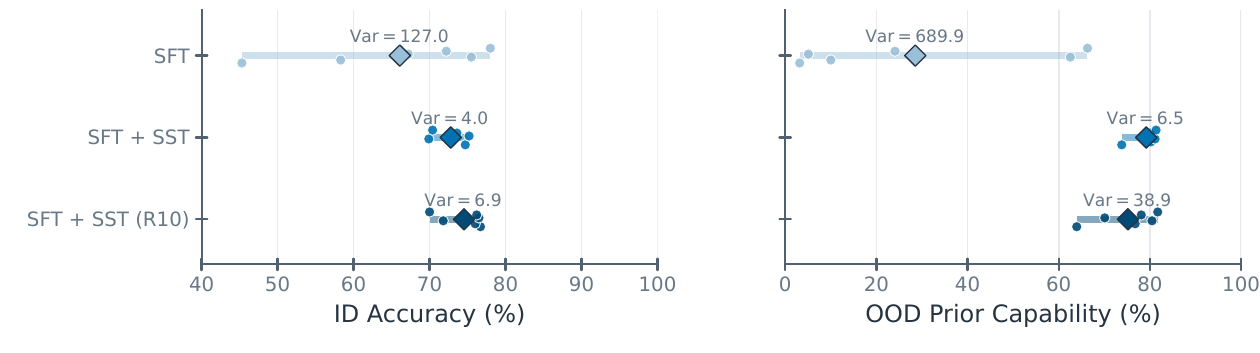}
\caption{\textbf{FinQA sensitivity to learning rate and schedule.}
Each dot is one of six configurations, the thick segment spans the minimum and maximum, and
the diamond marks the mean. SST greatly reduces variation in both in-domain
accuracy and the out-of-domain (OOD) prior-capability average.}
\label{fig:finqa-optimizer-hparams}
\end{figure}

We test sensitivity to two optimizer hyperparameters on FinQA: the learning
rate and its schedule. We cross three learning rates
($10^{-4}$, $5{\times}10^{-5}$, and $10^{-5}$) with flat and cosine schedules,
holding the data, training budget, and evaluation fixed. Figure~\ref{fig:finqa-optimizer-hparams}
shows all six configurations for SFT, SST without refresh, and SST with
refresh every 10 steps (R10), together with their minima, maxima, means, and
population variances.

SST is substantially more robust to these choices than unconstrained SFT. Both SST variants exceed SFT in-domain accuracy in four of
the six matched configurations while retaining prior capabilities. At the lowest learning rate, however, SFT
attains the highest individual in-domain scores, so this sweep does not imply
universal in-domain dominance. Rather, SST consistently improves retention and
generally yields a more stable joint learning--retention outcome; R10 shifts
the balance toward higher average in-domain accuracy at some cost to retention
robustness.

\subsection{Computational Cost}
\label{sec:cost}

In this section, we consider the computation cost of subspace-constrained finetuning. SST adds no extra trainable parameters or inference-time cost.
The method projects each optimizer update and optionally refreshes
the protected basis. Refresh uses Gram eigendecompositions, using the smaller Gram matrix for
rectangular weights. The multi-GPU training setup uses DeepSpeed ZeRO-3 \citep{deepspeed}, and
the estimates distribute the independent matrix operations across eight GPUs.

We report the measured SFT step from full-parameter bfloat16 Qwen3-8B training on eight NVIDIA L40S
GPUs, with effective batch size $16$, and
maximum sequence length $4096$. The measured SFT step took $1.73$ s. Across the model's $36$ layers
and $252$ projected matrices, projection adds 26.2 ms (1.5\%) per step in all settings. Refresh adds an amortized 62.4 ms per step for R100 and 623.5 ms for R10, giving total overheads of 5.1\% and 37.6\%, respectively. Splitting the SVD computation across the 8 GPUs brings the overhead to 0.6\% and 4.7\% respectively.

\subsection{Training and evaluation details}
\label{app:hparams}

\paragraph{Training.} Unless set by an official baseline recipe in
Appendix~\ref{app:baselines}, the
Qwen3 and Olmo studies use learning rate $5{\times}10^{-5}$ and effective batch
size $16$.. 
These settings are shared across paired recipe/SST comparisons and
were not selected separately for each method or domain. Training is
full-parameter except for the factored Sculpting and OPLoRA baselines and uses
DeepSpeed ZeRO-3 \citep{deepspeed}.

\paragraph{Datasets and synthetic traces.} The Qwen2.5 instruct study uses ToolAlpaca
\citep{toolalpaca}, HuatuoGPT-o1 \citep{huatuogpto1}, and SciKnowEval
\citep{sciknoweval}. Qwen3 Pathology and Medicine use MedMCQA questions and answer labels
\citep{medmcqa2022}. For each question, the generator receives the answer options followed by
\texttt{Choose the correct answer (A, B, C, or D) and explain your reasoning.} We use the
Bedrock model identifier \texttt{us.anthropic.claude-sonnet-4-6}; its response is serialized
inside \texttt{<think>}$\ldots$\texttt{</think>} with a boxed answer. The archived sets contain
$8{,}996$ Medicine traces and $8{,}999$ Pathology traces, all with a nonempty thinking block and
boxed answer.

\paragraph{Evaluation.} Evaluation uses vLLM \citep{vllm}. Prior capability is the average of
GPQA-Diamond, MMLU-Pro, GSM8K, HumanEval, and IFEval; in-domain accuracy uses the task's native
answer extractor.

For the Olmo-3 medical study, open-ended accuracy is graded with the Bedrock model identifier
\texttt{us.anthropic.claude-sonnet-4-6}. The judge receives the question, reference answer, and
full model response and returns \texttt{CORRECT} or \texttt{INCORRECT}. The fabrication metric is
a deterministic lower bound that flags explicit claims of an absent source, including
``reference solution,'' ``provided answer,'' and ``based on the feedback.'' None of the
$1{,}000$ evaluation prompts contains a trigger phrase.

\end{document}